\pdfoutput=1

\documentclass[11pt]{article}

\usepackage{EMNLP2023}

\usepackage{times}
\usepackage{latexsym}

\usepackage[T1]{fontenc}

\usepackage[utf8]{inputenc}
\usepackage{microtype}
\usepackage{inconsolata}

\usepackage{booktabs}
\usepackage{multirow}
\usepackage{amsmath}
\usepackage{amssymb}
\usepackage{graphicx}

\title{Analyzing Traditional and Neural
Approaches to Multilingual
Readability Assessment}

\author{
  Joshua Wong \\
  Harvard University \\
  \texttt{joshwong@harvard.edu}
  \And
  Chris Tanner\thanks{\hspace{1.3mm}Chris Tanner is also employed with Kensho Technologies in Cambridge, MA, USA} \\
  Massachusetts Institute of Technology \\
  \texttt{cwt@mit.edu}
}

\begin{document}
\maketitle
\begin{abstract}
Transformer-based models excel at Automatic Readability Assessment (ARA), yet feature-based models remain in active use because their predictions tie back to linguistic properties. This matters because readability labels are subjective and rater-dependent, so high accuracy on noisy ground truth may reflect surface patterns rather than the linguistic structure that defines difficulty. We test whether transformers internalize the same features as traditional models across Arabic, English, French, Hindi, and Russian using the \textsc{ReadMe++} dataset. Shapley Additive Explanations (SHAP) identify the features driving traditional classifiers, which we then use as TCAV concept sets to probe multilingual XLM-R and language-specific encoders. Transformers recover surface-length, syntactic, and lexical-diversity signals, and reflect the ordinal CEFR structure of the traditional models. Alignment varies by model family, language, and layer, with language-specific encoders tracking traditional models more clearly than XLM-R. High linear separability does not always imply directional influence, limiting linear probing for count-based readability features.
\end{abstract}

\section{Introduction}
\label{sec:introduction}
\begin{figure}[t]
    \centering
    \includegraphics[width=\columnwidth]{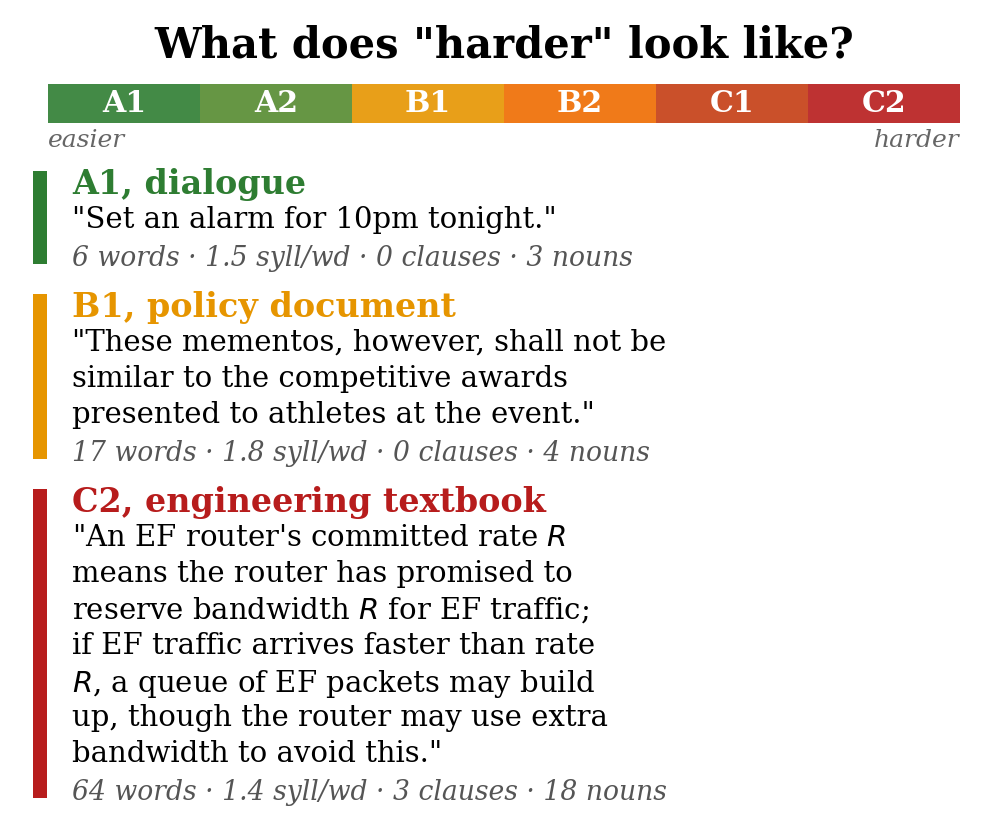}
    \caption{Sentences from \textsc{ReadMe++} across CEFR levels with four representative features: word count, syllables per word (vocabulary), subordinate clauses (syntactic depth), and noun count (information density).}
    \label{fig:readability_examples}
\end{figure}

Automatic Readability Assessment (ARA) predicts the difficulty of a text for a target reader. Difficulty is grounded in linguistic properties such as length, vocabulary, and syntactic structure, and these properties rise unevenly across reading levels as seen in Figure~\ref{fig:readability_examples}. Traditional feature-based classifiers remain in active use because their interpretability ties predictions back to these grounded linguistic features \citep{ribeiro-flucht_explainable_2024, imperial_under_2021}.
Throughout the paper, by traditional feature-based models, we refer to feature-based machine learning classifiers trained on handcrafted linguistic features, rather than to formula-based indices such as Flesch--Kincaid \citep{kincaid_derivation_1975}.
Transformer-based models achieve stronger performance but their internal representations are less transparent. Because readability labels are subjective and rater-dependent across small annotator pools \citep{naous_readme_2024, vajjala_trends_2022}, high accuracy on noisy ground truth may reflect surface patterns rather than the linguistic structure of difficulty, so accuracy alone is not a sufficient basis for trusting these models. Hybrid systems further suggest that explicit linguistic cues carry information transformers may not fully capture \citep{lee_pushing_2021, liu_hybrid_2023, wilkens_exploring_2024, imperial_bert_2021}. This raises a central question. Do transformer encoders learn the same readability-relevant linguistic signals that feature-based models rely on?

We investigate this question across Arabic, English, French, Hindi, and Russian using the \textsc{ReadMe++} dataset \citep{naous_readme_2024}. We first train traditional classifiers on handcrafted lexical, syntactic, and structural features, then apply Shapley Additive Explanations (SHAP) to identify which features drive their predictions \citep{lundberg_unified_2017}. We use these SHAP-selected predictors as concept sets for Testing with Concept Activation Vectors (TCAV), probing both multilingual XLM-R and language-specific transformer encoders \citep{kim_interpretability_2018}. This pairing lets us compare \emph{which features} drive traditional classifiers with \emph{which directions} influence neural models.

Some transformer encoders recover classical readability signals, including surface length, syntactic complexity, and lexical diversity. Some also reflect the ordinal structure of CEFR levels observed in feature-based models. This alignment varies across languages, layers, and model families, with language-specific encoders generally producing clearer TCAV patterns than XLM-R. High linear separability does not always imply directional influence for count-based features, limiting linear probing as evidence of functional concept use.

Our contributions are threefold. First, we compare feature-based and transformer-based readability classification models across five languages. Second, we characterize how linguistic feature alignment varies by language, model family, and encoder layer. Third, we introduce a SHAP--TCAV framework for testing whether neural readability models encode the same interpretable signals used by traditional ARA systems.

\section{Related Work}
\label{sec:related_work}

\paragraph{Feature-based, multilingual, and cross-lingual ARA.}
Early readability assessment relied on formula-based measures using surface features such as sentence length, word length, and unfamiliar word counts \citep{flesch_new_1948, kincaid_derivation_1975, dale_formula_1948, mc_laughlin_smog_1969}. Later machine learning approaches introduced richer handcrafted features, including lexical frequency, part-of-speech distributions, syntactic parse features, and discourse-level measures \citep{heilman_analysis_2008, vajjala_applicability_2013, deutsch_linguistic_2020}. These features remain useful because they connect predictions to interpretable linguistic properties. Uneven access to labeled data, readability standards, and NLP tools across languages limits multilingual ARA \citep{vajjala_trends_2022, rao_cross-lingual_2021}. Prior work addresses this through language-independent shallow features, language-specific models, and cross-lingual transfer using multilingual encoders \citep{shen_language-independent_2013, budd_using_2019, gupta_analyzing_2023, naous_readme_2024, imperial_universalcefr_2025}.
Most recently, UniversalCEFR consolidated existing CEFR corpora, including \textsc{ReadMe++}, and benchmarked feature-based models, fine-tuned encoders, and instruction-tuned LLMs across many languages and label granularities \citep{imperial_universalcefr_2025}. These studies evaluate predictive performance. They do not test whether neural models encode the linguistic features that feature-based ARA systems use, which is the question we address.

\paragraph{Transformer-based and hybrid ARA.}
Transformer encoders have become central to ARA, often outperforming traditional feature-based models \citep{martinc_supervised_2021, imperial_bert_2021, lee_pushing_2021}. Hybrid systems that combine transformer representations or predictions with handcrafted linguistic features often achieve strong results, especially in low-data or cross-domain settings \citep{lee_pushing_2021, liu_hybrid_2023, wilkens_exploring_2024}. This creates a tension. If explicit features improve transformer models, they may contain information not fully captured by neural representations. If they do not, transformers may already encode those signals internally \citep{deutsch_linguistic_2020}. Existing work addresses this question through downstream performance. We instead examine the representations directly, asking whether fine-tuned transformer readability models encode the same linguistic concepts that drive feature-based classifiers.

\paragraph{Feature attribution and concept-based probing.}
Interpretability methods provide tools for comparing explicit features and neural representations. SHAP assigns importance scores to input features and has been used to identify which linguistic predictors drive readability classifiers \citep{lundberg_unified_2017, ribeiro-flucht_explainable_2024, imperial_under_2021}.
Within readability itself, interpretability work remains limited. \citet{ribeiro-flucht_explainable_2024} explain feature-based CEFR classifiers with SHAP, \citet{wilkens_paying_2024} analyze which handcrafted features complement transformer ensembles, and \citet{sarti_that_2021} probe how fine-tuning on perceived-complexity prediction reshapes the linguistic features encoded in a neural language model, contrasting these with the features weighted by an SVM. However, this analysis targets perceived complexity rather than readability. Whether fine-tuned readability encoders rely on the same features that drive feature-based ARA classifiers remains untested.
Probing methods, by contrast, test whether linguistic properties can be recovered from neural representations \citep{conneau_what_2018, tenney_bert_2019}. High probing accuracy does not necessarily show that a model uses a concept for prediction. A feature may be linearly separable without being functionally influential \citep{maudslay_syntactic_2021, bolukbasi_interpretability_2021}. TCAV addresses this limitation by representing a human-defined concept as a direction in activation space and measuring whether movement along that direction affects model predictions \citep{kim_interpretability_2018}. Prior work applies TCAV to NLP classification tasks such as toxicity and abusive language detection \citep{nejadgholi_improving_2022, nejadgholi_towards_2022, nejadgholi_concept-based_2023}, but not to systematically compare feature-based and transformer-based ARA across languages. Our work fills this gap by pairing SHAP-derived readability features with TCAV-based concept probing.

\section{Methodology}
\label{sec:methodology}

We study whether transformer-based readability models encode the same linguistic signals used by traditional feature-based classifiers. Our pipeline has two stages. First, we train traditional classifiers on handcrafted linguistic features and use SHAP to identify the features most important for readability prediction. Second, we use those SHAP-selected features as TCAV concepts to probe multilingual and language-specific transformer encoders.

\paragraph{Data.}
We use \textsc{ReadMe++}, a multilingual sentence-level readability corpus covering Arabic, English, French, Hindi, and Russian \citep{naous_readme_2024}.
We use a single corpus so that all labels follow one annotation protocol at one granularity, and any alignment we observe reflects the models rather than shifting label conventions or mixed text lengths. \textsc{ReadMe++} meets this requirement while spanning 21 domains, and it targets L2 readability, whereas most alternative corpora target L1 native readers. UniversalCEFR offers broader language coverage but aggregates corpora with mixed granularities and uneven level distributions, and it already includes \textsc{ReadMe++} for our five languages \citep{imperial_universalcefr_2025}.
\textsc{ReadMe++} annotates sentences on the six-level CEFR scale (A1--C2), which we treat as a six-class classification task with classes indexed $k \in \{0,\ldots,5\}$ corresponding to A1 through C2. We use the publicly available portion of the dataset, containing 9{,}685 sentences, and create stratified train, validation, and test splits with a 60/10/30 ratio for each language. Table~\ref{tab:dataset_splits} reports the resulting split sizes. Class distributions are imbalanced and vary by language. English skews to A2--B2, Arabic to B1--B2, Russian to A1, French to B1, and Hindi is the most balanced. We use balanced class weights for all traditional classifiers.

\begin{table}[t]
\centering
\small
\setlength{\tabcolsep}{5pt}
\caption{Dataset sizes by language and split.}
\label{tab:dataset_splits}
\begin{tabular}{lrrrr}
\toprule
Language & Train & Val & Test & Total \\
\midrule
English  & 1{,}692 & 283 & 847 & 2{,}822 \\
Arabic   & 1{,}166 & 195 & 584 & 1{,}945 \\
French   & 1{,}001 & 167 & 501 & 1{,}669 \\
Hindi    &   894 & 149 & 448 & 1{,}491 \\
Russian  & 1{,}054 & 176 & 528 & 1{,}758 \\
\midrule
Total    & 5{,}807 & 970 & 2{,}908 & 9{,}685 \\
\bottomrule
\end{tabular}
\end{table}

For TCAV concept construction, we use the \textsc{ReadMe++} training split for all languages. For English and Arabic, we enlarge the concept pools with CEFR-SP \citep{arase_cefr-based_2022} and DARES \citep{el-haj_dares_2024}, respectively. The English pool combines the \textsc{ReadMe++} training split with the deduplicated Wiki-Auto and SCoRE subset of CEFR-SP, yielding roughly 12{,}000 sentences. The Arabic pool combines the \textsc{ReadMe++} training split with DARES, yielding roughly 15{,}000 sentences. We use these supplementary corpora only to construct concept and random sets, not for classifier training, validation, or testing.
These supplementary corpora only widen the spread of feature values available for concept construction. Because each CAV contrasts concept sentences with random sentences drawn from the same pool, a domain shift affects both sides equally rather than adding noise to one side. This follows the TCAV design of \citet{kim_interpretability_2018}, which defines concepts from example sets independent of the model's training data.

\paragraph{Linguistic features and feature selection.}
We extract handcrafted linguistic features using LFTK \citep{lee_lftk_2023}, with Stanza-based preprocessing through the \texttt{spacy-stanza} interface. LFTK extracts interpretable text properties such as surface length and density features, lexical diversity measures, part-of-speech composition, POS-based lexical variation, and entity-density features. For English, LFTK also provides additional English-specific features, including richer entity-density and lexical familiarity features based on external lexical resources. We exclude precomputed readability formula features so that the feature-based models do not directly encode the target construct. This yields 141 general LFTK features for non-English languages and 211 English features after excluding readability formulas.

Because many LFTK features are highly correlated, we apply correlation-based hierarchical clustering separately for each language on the training split. We first remove zero-variance features, compute pairwise Spearman correlations, define feature distance as $d(i,j)=1-|\rho_{ij}|$, and apply agglomerative clustering with complete linkage. We cut the dendrogram at $\tau=0.30$, corresponding to $|\rho|>0.70$ within a cluster, and keep one representative per cluster. The representative is the feature with the highest absolute Spearman correlation with the CEFR label. This reduces English from 211 to 59 features, and the non-English languages from 141 features to 32--35 selected features.

\paragraph{Traditional classifiers.}
We train three traditional classifiers per language on the selected linguistic features. The classifiers are Logistic Regression, Linear SVM, and Random Forest, following prior explainable ARA work \citep{ribeiro-flucht_explainable_2024}. We standardize features for Logistic Regression and Linear SVM but not for Random Forest, since tree-based models do not require scaling. All classifiers use balanced class weights. We tune hyperparameters with grid search and 10-fold stratified cross-validation, using Quadratic Weighted Kappa (QWK) as the scoring metric. QWK suits CEFR prediction because it penalizes larger level errors more strongly than adjacent errors. We then refit the best model for each language on the full training split before computing SHAP values.

\paragraph{Transformer models.}
We compare multilingual XLM-R base \citep{conneau_unsupervised_2020} with language-specific encoders. The language-specific models are BERT for English \citep{devlin_bert_2019}, AraBERTv02 for Arabic \citep{antoun_arabert_2020}, CamemBERT for French \citep{martin_camembert_2020}, MuRIL for Hindi \citep{khanuja_muril_2021}, and RuBERT for Russian \citep{kuratov_adaptation_2019}. We fine-tune each model separately per language for six-way CEFR classification, following \citet{naous_readme_2024}.
We train for 20 epochs with cross-entropy loss and Adam, select the learning rate from $\{10^{-5},10^{-6},10^{-7}\}$ by validation QWK, and use the best validation-QWK checkpoint. The search selects $10^{-5}$ for all languages. Each model uses a fresh six-way classification head on the \texttt{[CLS]} token representation and has 12 encoder layers, which we probe with TCAV.

\paragraph{SHAP-based feature ranking.}
We use SHAP to identify which features drive traditional classifiers \citep{lundberg_unified_2017}. We choose SHAP over alternatives such as LIME because it satisfies local accuracy and consistency, which LIME can violate \citep{lundberg_unified_2017, zhao_explainability_2023}. Consistency also makes importance comparable across our three classifier families, which the Borda aggregation below requires.
We use model-specific explainers that compute Shapley values exactly for each model class. For Logistic Regression and Linear SVM, we use \texttt{LinearExplainer} with correlation-dependent perturbation, using the training split as the background distribution. This estimates missing features conditionally on observed features rather than assuming feature independence, which accounts for feature correlations that clustering mitigated. For Random Forest, we use \texttt{TreeExplainer}, which computes SHAP values directly from the tree structure.

For each feature, we compute global importance as the mean absolute SHAP value across test examples and CEFR classes:
\[
\mathrm{Imp}_m(i) =
\frac{1}{nK}
\sum_{j=1}^{n}
\sum_{k=1}^{K}
|S^{m}_{j,i,k}|,
\]
where $S^{m}_{j,i,k}$ is the SHAP value of feature $i$ for example $j$ and class $k$ under model $m$.

The explainers produce SHAP values on different scales. Therefore, we rank features within each classifier and aggregate the ranks using Borda count:
\[
B(i)=\frac{1}{3}\sum_{m} r_m(i),
\]
where $r_m(i)$ is the rank of feature $i$ under model $m$. For each language, we select the 10 features with the lowest Borda scores and the 10 features with the highest Borda scores as TCAV concepts.

\paragraph{TCAV concept construction and probing.}
We use TCAV to test whether SHAP-selected readability features correspond to influential directions in transformer representations \citep{kim_interpretability_2018}. We choose TCAV among concept-based probing methods because it yields semantically coherent concepts \citep{li_evaluating_2024} and because it measures directional influence on predictions rather than linear recoverability alone.

For each selected feature, the positive concept set consists of 100 sentences with the highest values for that feature in the concept pool. Each concept therefore represents the property of having a high value of that feature: the concept for \texttt{t\_char} is sentences with many characters, and the concept for \texttt{t\_syll2} is sentences with many words of more than two syllables. TCAV then tests whether moving along this high-feature direction in activation space affects the model's CEFR predictions. Each random set contains 100 sentences sampled from the same pool. We draw 1{,}000 random sets per concept for English and Arabic, where the concept pools are enlarged with CEFR-SP and DARES, and 500 random sets per concept for French, Hindi, and Russian, where concept pools are limited to the \textsc{ReadMe++} training split.

For each transformer layer $l$, we extract the \texttt{[CLS]} representation $h_l(x)$ and train a logistic regression classifier to distinguish concept activation from random activations. The normalized classifier weight vector defines the Concept Activation Vector (CAV):
\[
v_C^{(l)}=\frac{w_C^{(l)}}{\|w_C^{(l)}\|_2}.
\]
We record CAV classifier accuracy as a measure of linear separability. We then measure whether moving in the concept direction increases the model's predicted probability for CEFR class $k$:
\[
D_{C,k,l}(x)=\nabla_{h_l}p(y=k \mid x)\cdot v_C^{(l)}.
\]
A positive value means moving the hidden representation along the concept direction raises the predicted probability for class $k$. The TCAV score is the fraction of class-$k$ test examples with a positive directional derivative:
\[
\mathrm{TCAV}_{C,k,l}
=
\frac{1}{|X_k|}
\sum_{x\in X_k}
\mathbf{1}\left[D_{C,k,l}(x)>0\right].
\]
CAV accuracy measures whether a concept is linearly recoverable from a layer. The TCAV score measures whether that concept direction positively influences the model's prediction for a CEFR class.

For probing, we extract activations and gradients at a maximum sequence length of 512 tokens, the architectural input limit of every target encoder. This recovers full context for the long-sentence tail. We compute gradients on the held-out test set, partitioned by true CEFR level, so each per-class pool $X_k$ contains only test examples with the corresponding rating.

\paragraph{TCAV significance testing.}
Each concept is evaluated across multiple random sets, producing a distribution of TCAV scores for each concept--class--layer tuple $(C,k,l)$. We test whether the mean TCAV score differs from the random baseline of 0.5 using a two-sided one-sample $t$-test:
\[
H_0: \mu_{C,k,l}=0.5.
\]
We assess significance at $\alpha=0.01$ with Bonferroni correction. For each plotted analysis, we divide $\alpha$ by the maximum number of concept--class--layer comparisons shown in the corresponding graph and apply the adjusted threshold uniformly across plots. A TCAV result is significant only if its $p$-value falls below this Bonferroni-adjusted threshold.

\section{Results}

\subsection{Model Performance}

Table~\ref{tab:performance} reports test performance. Both transformer families outperform the traditional classifiers on QWK and accuracy across every language, with the language-specific encoder doing so on macro F1 as well in English, Arabic, and Russian. The improvement is largest on ordinal metrics. QWK ranges from 0.65--0.78 for the best traditional classifiers up to 0.83--0.86 for the best transformers. Accuracy improves from around 0.44--0.51 up to 0.53--0.64.

The two transformer families perform at broadly similar levels on QWK, but diverge more on accuracy and macro F1. XLM-R and BERT are nearly tied in English, and XLM-R and RuBERT are nearly tied in Russian, with RuBERT slightly higher on accuracy and XLM-R higher on macro F1. AraBERTv02 shows the clearest language-specific advantage, gaining 6 accuracy points and 6 macro F1 points over XLM-R in Arabic. The pattern reverses in Hindi and French, where XLM-R outperforms MuRIL and CamemBERT on accuracy and macro F1 despite similar QWK. The macro F1 gap is largest in French (0.53 versus 0.36) and Hindi (0.54 versus 0.41), likely from the balanced Hindi class distribution and CamemBERT's known sensitivity on smaller fine-tuning sets, where underperformance on any single class hurts the macro average more than in skewed languages.

\begin{table}[t]
\centering
\footnotesize
\caption{Test QWK, Accuracy, and Macro F1 per model and language. RF = Random Forest. LS = language-specific encoder. Best in bold.}
\label{tab:performance}
\begin{tabular}{@{}lccccc@{}}
\toprule
Model & EN & AR & FR & HI & RU \\
\midrule
\multicolumn{6}{c}{\textit{QWK}} \\
LR & 0.71 & 0.60 & 0.74 & 0.76 & 0.75 \\
SVM & 0.72 & 0.65 & 0.75 & 0.77 & 0.76 \\
RF & 0.75 & 0.65 & 0.78 & 0.76 & 0.76 \\
XLM-R & 0.82 & 0.82 & \textbf{0.83} & \textbf{0.83} & \textbf{0.86} \\
LS & \textbf{0.83} & \textbf{0.85} & 0.81 & 0.82 & \textbf{0.86} \\
\midrule
\multicolumn{6}{c}{\textit{Accuracy}} \\
LR & 0.40 & 0.43 & 0.40 & 0.44 & 0.42 \\
SVM & 0.45 & 0.44 & 0.40 & 0.44 & 0.42 \\
RF & 0.51 & 0.50 & 0.46 & 0.45 & 0.44 \\
XLM-R & 0.59 & 0.58 & \textbf{0.53} & \textbf{0.54} & 0.55 \\
LS & \textbf{0.60} & \textbf{0.64} & 0.51 & 0.46 & \textbf{0.56} \\
\midrule
\multicolumn{6}{c}{\textit{Macro F1}} \\
LR & 0.37 & 0.43 & 0.38 & 0.45 & 0.40 \\
SVM & 0.41 & 0.45 & 0.39 & 0.45 & 0.41 \\
RF & 0.46 & 0.49 & 0.44 & 0.45 & 0.38 \\
XLM-R & 0.50 & 0.59 & \textbf{0.53} & \textbf{0.54} & \textbf{0.51} \\
LS & \textbf{0.54} & \textbf{0.65} & 0.36 & 0.41 & 0.47 \\
\bottomrule
\end{tabular}
\end{table}

\subsection{What Feature-Based Models Use}

Table~\ref{tab:feature_descriptions} gives the full name and a short description for every feature that appears in any language's top 10. Surface-length features fill the top three positions in every language. Total characters (\texttt{t\_char}) appears in the top three for every language, and average characters per word (\texttt{a\_char\_pw}) appears in the top four for every language (top three in English, Arabic, Hindi, and Russian, and rank four in French). Type-token ratio (\texttt{simp\_ttr}) appears in the top 7 for all five, and the length-normalized Uber Index (\texttt{uber\_ttr}) enters the top 10 for Arabic, Hindi, and Russian. POS counts fill the remaining slots. Adjectives (\texttt{n\_adj}) are high-ranked in Arabic, French, Hindi, and Russian, and adpositions (\texttt{n\_adp}) are high-ranked in English, French, Hindi, and Russian. English diverges mainly through two features only available in its enlarged pool. \texttt{t\_syll2} counts multi-syllabic words, and \texttt{a\_kup\_pw} measures Kuperman age-of-acquisition per word.

\begin{table}[t]
\centering
\scriptsize
\setlength{\tabcolsep}{3pt}
\renewcommand{\arraystretch}{1.05}
\caption{Descriptions of all features appearing in any language's top 10. The Abbr.\ column gives the short label used in Table~\ref{tab:features}. Group abbreviations: SLD = surface length and density, POS = POS composition, LV = POS-based lexical variation, LD = lexical diversity, LF = lexical familiarity.}
\label{tab:feature_descriptions}
\begin{tabular*}{\columnwidth}{@{\extracolsep{\fill}}lllp{3.5cm}@{}}
\toprule
Feature & Abbr. & Group & Description \\
\midrule
\texttt{t\_char}          & char     & SLD & Total characters in the sentence \\
\texttt{t\_syll2}         & syll2    & SLD & Words with more than two syllables (English only) \\
\texttt{t\_stopword}      & stop     & SLD & Total stop words \\
\texttt{a\_char\_ps}      & char/s   & SLD & Avg characters per sentence \\
\texttt{a\_char\_pw}      & char/w   & SLD & Avg characters per word \\
\midrule
\texttt{n\_noun}          & noun     & POS & Count of nouns \\
\texttt{n\_adj}           & adj      & POS & Count of adjectives \\
\texttt{n\_adp}           & adp      & POS & Count of adpositions \\
\texttt{n\_det}           & det      & POS & Count of determiners \\
\texttt{n\_cconj}         & cconj    & POS & Count of coord.\ conjunctions \\
\texttt{n\_ucconj}        & uconj    & POS & Unique coord.\ conjunctions \\
\texttt{n\_udet}          & udet     & POS & Unique determiners \\
\texttt{n\_uadv}          & uadv     & POS & Unique adverbs \\
\texttt{n\_part}          & part     & POS & Count of particles \\
\texttt{a\_verb\_pw}      & verb/w   & POS & Avg verbs per word \\
\texttt{a\_verb\_ps}      & verb/s   & POS & Avg verbs per sentence \\
\texttt{a\_noun\_pw}      & noun/w   & POS & Avg nouns per word \\
\texttt{a\_pron\_pw}      & pron/w   & POS & Avg pronouns per word \\
\texttt{a\_punct\_ps}     & punct/s  & POS & Avg punctuation per sentence \\
\midrule
\texttt{simp\_adj\_var}   & adj-var  & LV  & Unique adj.\ lemmas / adj.\ tokens \\
\texttt{simp\_cconj\_var} & cconj-var & LV & Unique cconj lemmas / cconj tokens \\
\texttt{root\_adp\_var}   & adp-var  & LV  & Unique adp lemmas / $\sqrt{\text{adp tokens}}$ \\
\texttt{root\_verb\_var}  & verb-var & LV  & Unique verb lemmas / $\sqrt{\text{verb tokens}}$ \\
\midrule
\texttt{simp\_ttr}        & TTR      & LD  & Type-token ratio: unique / total \\
\texttt{uber\_ttr}        & Uber     & LD  & Uber Index: $\log^{2}(\text{total}) / \log(\text{total}/\text{unique})$ \\
\midrule
\texttt{a\_kup\_pw}       & AoA/w    & LF  & Avg Kuperman age-of-acquisition per word (English only) \\
\bottomrule
\end{tabular*}
\end{table}

\begin{figure}[t]
    \centering
    \includegraphics[width=\columnwidth]{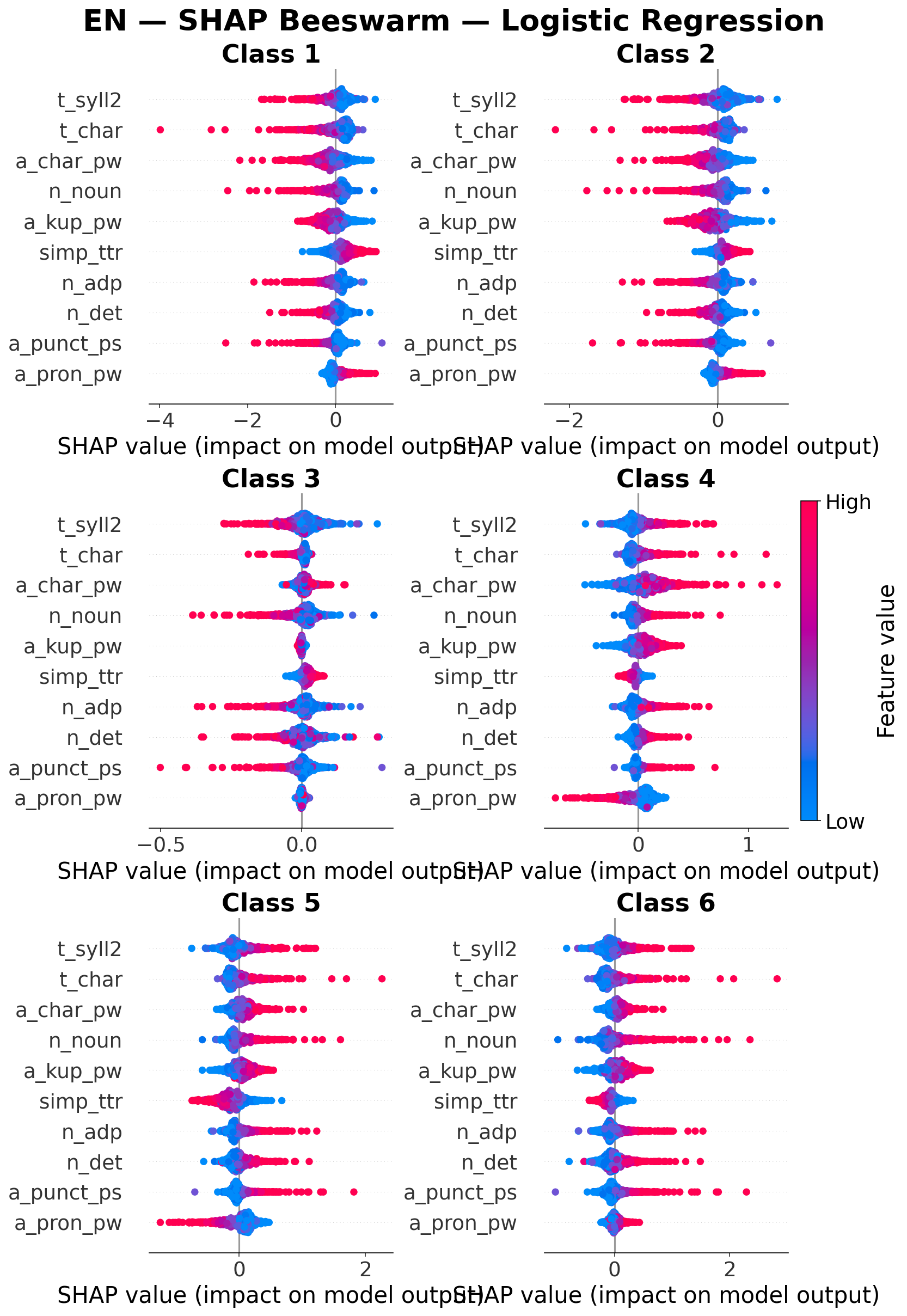}
    \caption{English Logistic Regression SHAP beeswarm, one panel per CEFR level (Class 0 = A1, Class 5 = C2). Color encodes feature value (blue = low, red = high), and horizontal position encodes SHAP contribution. Surface-length features show inverted polarity between A1 and C2, indicating an ordinally coherent representation despite no ordinal loss.}
    \label{fig:shap_beeswarm}
\end{figure}

Per-class SHAP beeswarms show that features invert polarity cleanly between the extreme CEFR classes (Figure~\ref{fig:shap_beeswarm}). For A1 (Class 0), high values of \texttt{t\_char}, \texttt{a\_char\_pw}, and related count features produce negative SHAP contributions. For C2 (Class 5), the same high values produce positive contributions. Intermediate classes transition smoothly between these poles. Classes 1 and 2 share the A1 polarity with decreasing magnitude. Classes 3 and 4 share the C2 polarity with increasing magnitude. Class 2 shows the most compressed range, marking the transition between the two regimes. This ordinal coherence is notable because the classifiers use flat cross-entropy with no ordinal constraint. The same pattern holds for the SVM and Random Forest classifiers and for the other four languages.

Table~\ref{tab:features} shows the top 10 features by Borda rank per language, with the direction in which each feature pushes predictions. Feature directions match linguistic intuition. Longer sentences and longer words push toward harder CEFR levels, consistent with classical readability formulas. More adjectives, adpositions, and conjunctions push toward harder levels, indexing modification, prepositional phrases, and clause coordination. Raw \texttt{simp\_ttr} decreases mechanically with sentence length and pushes toward easier text. Length-normalized \texttt{uber\_ttr} pushes the other way in Arabic, Hindi, and Russian, capturing genuine vocabulary richness once length is controlled for. For English, \texttt{a\_kup\_pw} pushes upward, meaning sentences built from later-learned words rate as harder, and \texttt{a\_pron\_pw} trends downward, since pronoun-heavy text is typically conversational and easier.

\begin{table}[t]
\centering
\scriptsize
\setlength{\tabcolsep}{2pt}
\renewcommand{\arraystretch}{0.92}
\caption{Top 10 features per language by Borda rank, with Logistic Regression SHAP direction. Abbreviations defined in Table~\ref{tab:feature_descriptions}.}
\label{tab:features}
\begin{tabular*}{\columnwidth}{@{\extracolsep{\fill}}rlllll@{}}
\toprule
Rank & EN & AR & FR & HI & RU \\
\midrule
1  & syll2$\uparrow$    & char$\uparrow$      & char/s$\uparrow$  & char/w$\uparrow$  & char$\uparrow$ \\
2  & char$\uparrow$     & char/w$\uparrow$    & char$\uparrow$    & char$\uparrow$    & char/w$\uparrow$ \\
3  & char/w$\uparrow$   & stop$\uparrow$      & TTR$\downarrow$   & noun$\uparrow$    & adj$\uparrow$ \\
4  & noun$\uparrow$     & adj$\uparrow$       & char/w$\uparrow$  & adp$\uparrow$     & noun/w$\uparrow$ \\
5  & AoA/w$\uparrow$    & uconj$\uparrow$     & adp$\uparrow$     & cconj$\uparrow$   & Uber$\uparrow$ \\
6  & TTR$\downarrow$    & TTR$\downarrow$     & adj$\uparrow$     & adj$\uparrow$     & adp$\uparrow$ \\
7  & adp$\uparrow$      & part$\uparrow$      & udet$\uparrow$    & TTR$\downarrow$   & TTR$\downarrow$ \\
8  & det$\uparrow$      & verb/w$\downarrow^{*}$ & verb/s$\uparrow$ & adj-var$\uparrow$ & stop$\uparrow$ \\
9  & punct/s$\uparrow$  & Uber$\uparrow^{*}$  & uadv$\uparrow$    & Uber$\uparrow$    & cconj$\uparrow$ \\
10 & pron/w$\downarrow^{*}$ & cconj-var$\uparrow^{*}$ & punct/s$\uparrow$ & adp-var$\uparrow$ & verb-var$\uparrow$ \\
\bottomrule
\end{tabular*}

\vspace{1mm}
\begin{minipage}{\columnwidth}
\scriptsize
\textit{Note.} $\uparrow$ means higher feature values push toward higher CEFR levels. $\downarrow$ means higher values push toward lower CEFR levels. $^{*}$ marks a general trend not consistent across all class-level panels.
\end{minipage}
\end{table}

\subsection{CAV Separability}
\label{sec:cav_separability}

Across all five languages, CAVs trained on the top 10 SHAP features reach higher classifier accuracy than CAVs for the bottom 10.

\begin{figure}[ht]
    \centering
    \includegraphics[width=\columnwidth]{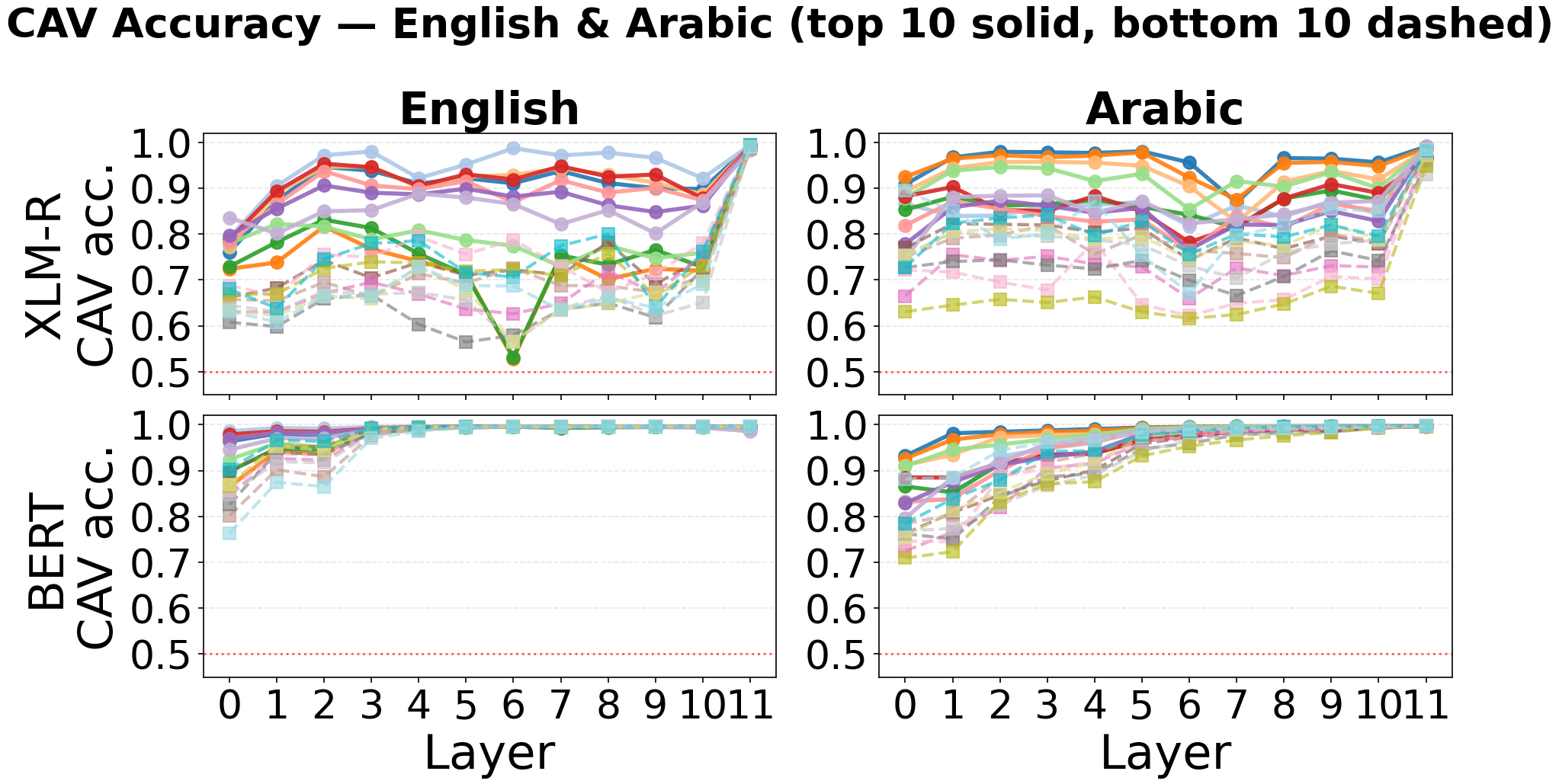}
    \caption{Mean CAV classifier accuracy across 12 layers for English and Arabic. Top row: XLM-R. Bottom row: language-specific encoder. Solid lines show top 10 SHAP features; dashed lines show bottom 10.}
    \label{fig:cav_acc}
\end{figure}

Figure~\ref{fig:cav_acc} shows CAV accuracies for Arabic and English.
The gap is visible for both transformer families. In the language-specific encoders, top-feature CAV accuracy generally rises in early layers. English BERT and Arabic AraBERTv02 reach near-perfect separability by mid-network and plateau, and Russian RuBERT reaches its plateau by layer 4. French CamemBERT and Hindi MuRIL are non-monotonic, with MuRIL peaking in early layers and dipping through the middle and final layers. In XLM-R, accuracy is noisier across layers. English XLM-R shows a dip around layer 6 before recovering at later layers. Hindi XLM-R is flatter across layers, while Hindi MuRIL peaks higher in early layers but dips in mid and final layers. The features driving the feature-based classifiers are the same features most linearly distinguishable in transformer activation spaces.

\subsection{TCAV Directionality}

The SHAP-derived directions transfer to the transformers in four of five languages. Concepts built from length, part-of-speech, and diversity features push predictions toward C2 and away from A1, and type-token ratio inverts, matching its SHAP polarity. The coherent layer groups differ by language and model family: English BERT at layers 0--5, Russian RuBERT at layers 4--6, French CamemBERT at layers 6--11, and Hindi XLM-R at layers 3--5. Arabic is the exception, with no consistent directionality in either family. The rest of this section walks through these cases.

\begin{figure}[t]
    \centering
    \includegraphics[width=\columnwidth]{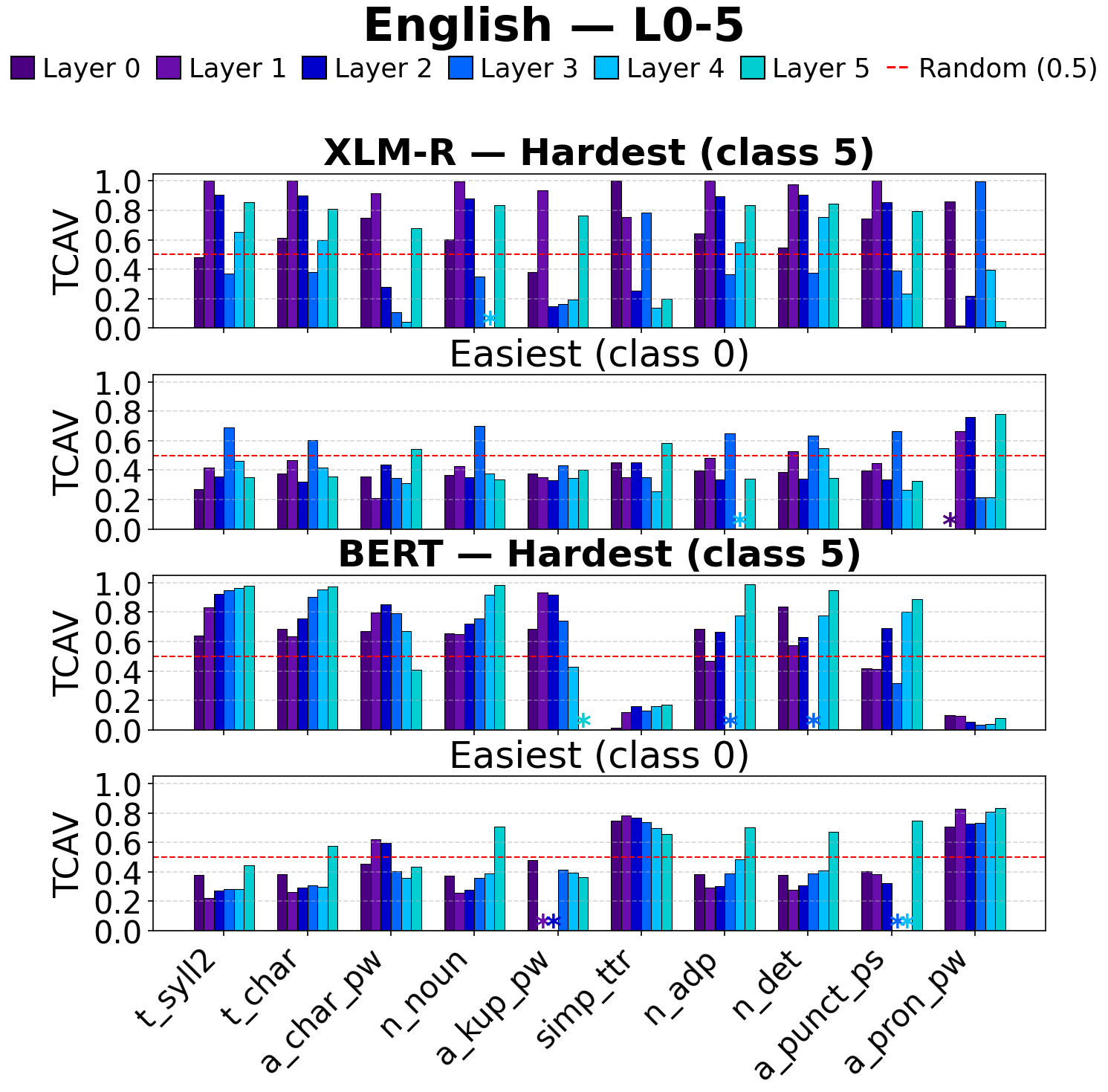}
    \caption{English TCAV fractions for the top 10 SHAP features, layers 0--5. For each model, the upper panel shows C2 (hardest) and the lower panel shows A1 (easiest). Asterisks mark non-significant results.}
    \label{fig:tcav_en}
\end{figure}

For four of the five languages, TCAV fractions in at least one model family reproduce the SHAP-derived directions, with class C2 fractions for count-based features sitting above 0.5 and class A1 fractions sitting below. Rather than reporting individual layers, we look for sequential layer groups where the directional pattern holds for most top-10 features, since a signal that appears at one layer but disappears at the next is hard to interpret as a stable property of the model. The cleanest example is English BERT (Figure~\ref{fig:tcav_en}, lower half). Across layers 0--5, count-based features (\texttt{t\_syll2}, \texttt{t\_char}, \texttt{a\_char\_pw}, \texttt{n\_noun}, \texttt{n\_adp}, \texttt{n\_det}, \texttt{a\_kup\_pw}, \texttt{a\_punct\_ps}) show high TCAV for C2 and low TCAV for A1. Moving the representation toward longer, denser sentences increases the model's C2 probability and decreases its A1 probability. Two features invert. \texttt{simp\_ttr} and \texttt{a\_pron\_pw} show low TCAV for C2 and high TCAV for A1, matching their SHAP polarity.

English XLM-R (Figure~\ref{fig:tcav_en}, upper half) shares almost none of this consistency. Several features swing widely across adjacent layers, and this instability lines up with the mid-layer dip in CAV accuracy from Section~\ref{sec:cav_separability}, suggesting XLM-R's intermediate representations encode these concepts less reliably. The same broad pattern holds for Russian and French. RuBERT holds the directional split across layers 4--6 and CamemBERT builds it progressively across layers 6--11, while XLM-R is unstable across the same ranges in both languages. The pattern holds only within these mid-to-late groups, not across all 12 layers.

\begin{figure}[t]
    \centering
    \includegraphics[width=\columnwidth]{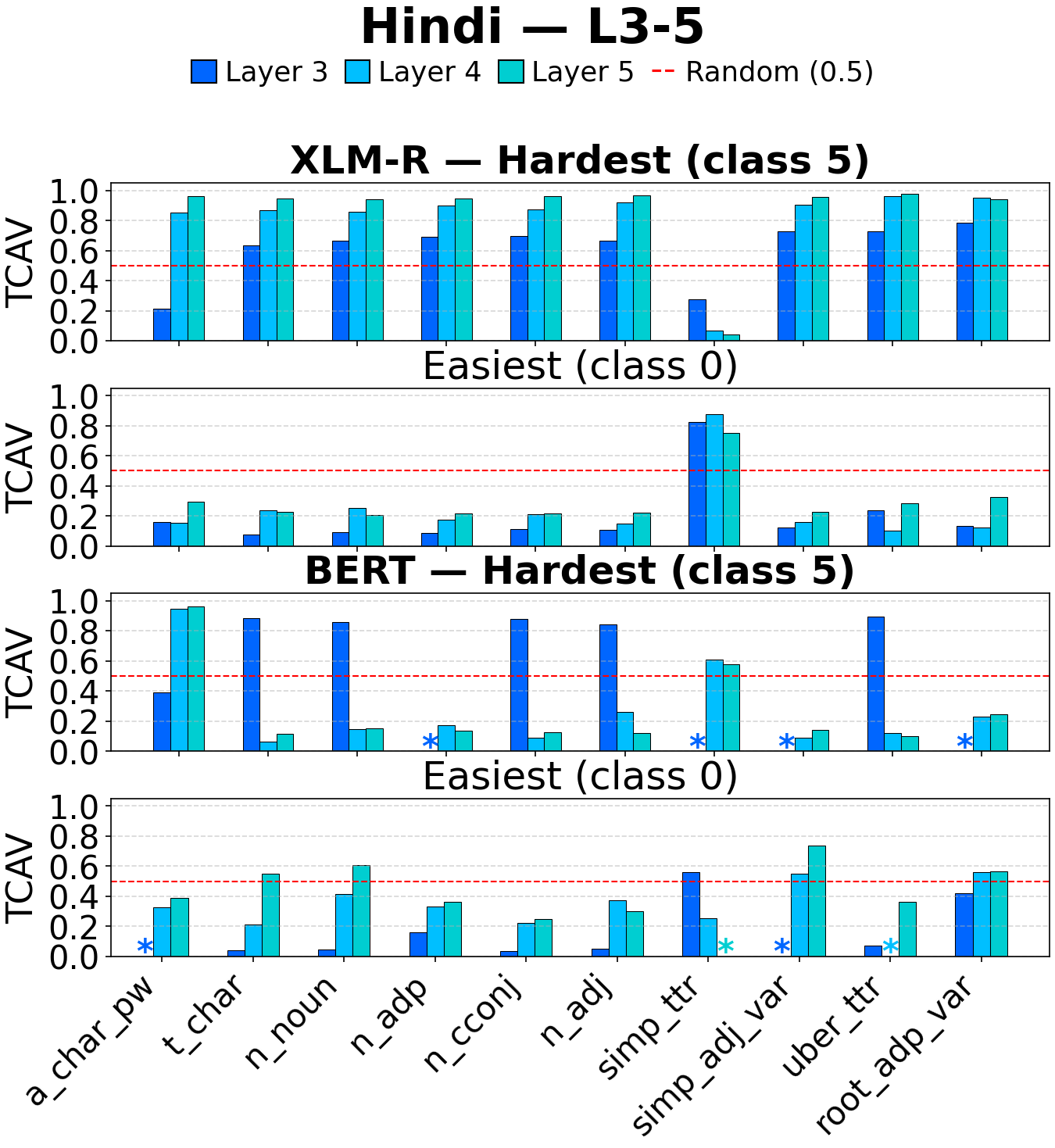}
    \caption{Hindi TCAV fractions for the top 10 SHAP features, layers 3--5. For each model, the upper panel shows C2 (hardest) and the lower panel shows A1 (easiest). Asterisks mark non-significant results.}
    \label{fig:tcav_hi}
\end{figure}

Hindi reverses the model-family pattern (Figure~\ref{fig:tcav_hi}). XLM-R produces the cleaner pattern across layers 3--5, with count features high for C2, \texttt{simp\_ttr} inverted, and almost all A1 fractions below 0.5. This is the only language where XLM-R produces clearer directional structure than the language-specific encoder. Section~\ref{sec:discussion} considers possible reasons.

Arabic breaks the pattern entirely. CAV accuracies are high for both transformer families, with top features reaching 0.90--0.98 in Figure~\ref{fig:cav_acc}, so the concepts are separable in activation space. The TCAV fractions for C2 and A1, however, do not consistently separate above and below the 0.5 baseline within any layer group we examined, for either AraBERTv02 or XLM-R. The concepts are linearly recoverable but do not act as the axes along which Arabic predictions move. We weigh candidate explanations for this gap in Section~\ref{sec:discussion}.

\section{Discussion}
\label{sec:discussion}

The feature-based classifiers and fine-tuned transformers converge on the same signals and the same directions. We defined the TCAV concepts from the feature-based side and applied them to transformers that never saw feature values as inputs. The agreement in both separability and polarity therefore indicates that the transformers rediscover this structure rather than encoding something categorically different.

How clearly the agreement surfaces depends on language and model family. Language-specific encoders give stable layer-wise trajectories while XLM-R is noisier in the middle layers. Hindi reverses this pattern, and Arabic departs from it in a different way. Our design cannot isolate the causes, so we state the explanations below as hypotheses rather than conclusions. 

We attribute the Hindi reversal to the data rather than to the language. Hindi is our smallest subset and has the most balanced class distribution, which plausibly makes CAV training less stable for the language-specific encoder. MuRIL's limited and partly transliterated pretraining data may add to this, while XLM-R's broader multilingual pretraining keeps its middle layers cleaner. Varying training-set size and class balance would test this hypothesis directly.

Arabic shows a precise gap rather than a failure. CAV accuracy reaches 0.90 to 0.98 for both AraBERTv02 and XLM-R, so the concepts are linearly separable, yet the TCAV fractions do not separate consistently for either family. The signals sit in activation space without acting as the axes along which predictions move. 
Our hypothesis is that Arabic root-and-pattern morphology may weaken the linear link between surface counts and difficulty, relative to Hindi and Russian, which kept directional structure with \textsc{ReadMe++} pools alone. Morphology-aware concepts and non-linear probes are the natural next test.

One task-level property cuts across both analyses. The feature-based beeswarms grade smoothly across all six CEFR levels, and the transformer TCAV results show the same inversion between A1 and C2 at the extremes, despite flat cross-entropy training in both cases. We designed neither analysis to produce ordinal structure, so the ordinal shape comes from the task itself. This pattern hints that explicit ordinal losses may add less than sometimes assumed. Testing that possibility requires fine-tuning with an ordinal objective, which we leave to future work.

\section{Conclusion}
\label{conclusion}

We read these results as validation. The performance gains of transformer ARA rest on linguistic features the field already trusts, which places the shift toward transformer encoders on firmer interpretive footing than performance numbers alone. The SHAP--TCAV pairing supplies independent evidence, explicit feature usage on one side and representation geometry on the other, so agreement narrows the space of explanations and disagreement, as in Arabic, locates the limit precisely. The procedure transfers in principle to any setting where an interpretable baseline and a neural model coexist, and broader validation remains future work.

\section*{Limitations}

The experiments use a single dataset (\textsc{ReadMe++}) at the sentence level on the CEFR scale, and we do not average transformer results across random seeds. The concept vocabulary is cross-lingually asymmetric because general-purpose LFTK features are narrower for non-English languages, particularly for entity density and lexical familiarity. TCAV tests directional influence along pre-specified linear directions, cannot reveal concepts the investigator did not specify, and is sensitive to random-set choice and concept entanglement. The comparison covers one multilingual and one language-specific encoder per language and does not generalize to larger or differently structured models. Natural next steps are a richer concept vocabulary for morphologically rich languages, causal methods such as activation patching or concept steering, and extension to instruction-tuned LLMs and document-level ARA.

\section*{Ethics Statement}
This work uses publicly available readability corpora and does not involve human subjects beyond the existing annotations in those corpora. All data sources are cited, and all fine-tuned models are derived from publicly released checkpoints under permissive licenses. The methods studied here are interpretability tools for classification models and do not introduce direct misuse risks. One indirect consideration is that readability models trained on limited-domain data may produce biased judgments when deployed in educational settings. The cross-lingual analysis highlights uneven feature availability across languages, which is itself an equity concern for multilingual ARA.

\bibliography{anthology,references}

\begin{thebibliography}{43}
\expandafter\ifx\csname natexlab\endcsname\relax\def\natexlab#1{#1}\fi

\bibitem[{Antoun et~al.(2020)Antoun, Baly, and Hajj}]{antoun_arabert_2020}
Wissam Antoun, Fady Baly, and Hazem Hajj. 2020.
\newblock \href {https://aclanthology.org/2020.osact-1.2/} {{AraBERT}:
  {Transformer}-based {Model} for {Arabic} {Language} {Understanding}}.
\newblock In \emph{Proceedings of the 4th {Workshop} on {Open}-{Source}
  {Arabic} {Corpora} and {Processing} {Tools}, with a {Shared} {Task} on
  {Offensive} {Language} {Detection}}, pages 9--15, Marseille, France. European
  Language Resources Association.

\bibitem[{Arase et~al.(2022)Arase, Uchida, and
  Kajiwara}]{arase_cefr-based_2022}
Yuki Arase, Satoru Uchida, and Tomoyuki Kajiwara. 2022.
\newblock \href {https://doi.org/10.18653/v1/2022.emnlp-main.416}
  {{CEFR}-{Based} {Sentence} {Difficulty} {Annotation} and {Assessment}}.
\newblock In \emph{Proceedings of the 2022 {Conference} on {Empirical}
  {Methods} in {Natural} {Language} {Processing}}, pages 6206--6219, Abu Dhabi,
  United Arab Emirates. Association for Computational Linguistics.

\bibitem[{Bolukbasi et~al.(2021)Bolukbasi, Pearce, Yuan, Coenen, Reif, Viégas,
  and Wattenberg}]{bolukbasi_interpretability_2021}
Tolga Bolukbasi, Adam Pearce, Ann Yuan, Andy Coenen, Emily Reif, Fernanda
  Viégas, and Martin Wattenberg. 2021.
\newblock \href {https://doi.org/10.48550/arXiv.2104.07143} {An
  {Interpretability} {Illusion} for {BERT}}.
\newblock ArXiv:2104.07143 [cs].

\bibitem[{Budd et~al.(2019)Budd, Marius, Gatewood, and Jones}]{budd_using_2019}
Ray Budd, Tamas Marius, Paul Gatewood, and Doug Jones. 2019.
\newblock \href {https://doi.org/10.21437/SLaTE.2019-16} {Using {K}-{Means} in
  {SVR}-{Based} {Text} {Difficulty} {Estimation}}.
\newblock In \emph{8th {ISCA} {Workshop} on {Speech} and {Language}
  {Technology} in {Education} ({SLaTE} 2019)}, pages 84--88. ISCA.

\bibitem[{Conneau et~al.(2020)Conneau, Khandelwal, Goyal, Chaudhary, Wenzek,
  Guzmán, Grave, Ott, Zettlemoyer, and Stoyanov}]{conneau_unsupervised_2020}
Alexis Conneau, Kartikay Khandelwal, Naman Goyal, Vishrav Chaudhary, Guillaume
  Wenzek, Francisco Guzmán, Edouard Grave, Myle Ott, Luke Zettlemoyer, and
  Veselin Stoyanov. 2020.
\newblock \href {https://doi.org/10.18653/v1/2020.acl-main.747} {Unsupervised
  {Cross}-lingual {Representation} {Learning} at {Scale}}.
\newblock In \emph{Proceedings of the 58th {Annual} {Meeting} of the
  {Association} for {Computational} {Linguistics}}, pages 8440--8451, Online.
  Association for Computational Linguistics.

\bibitem[{Conneau et~al.(2018)Conneau, Kruszewski, Lample, Barrault, and
  Baroni}]{conneau_what_2018}
Alexis Conneau, German Kruszewski, Guillaume Lample, Loïc Barrault, and Marco
  Baroni. 2018.
\newblock \href {https://doi.org/10.18653/v1/P18-1198} {What you can cram into
  a single \$\&!\#* vector: {Probing} sentence embeddings for linguistic
  properties}.
\newblock In \emph{Proceedings of the 56th {Annual} {Meeting} of the
  {Association} for {Computational} {Linguistics} ({Volume} 1: {Long}
  {Papers})}, pages 2126--2136, Melbourne, Australia. Association for
  Computational Linguistics.

\bibitem[{Dale and Chall(1948)}]{dale_formula_1948}
Edgar Dale and Jeanne~S. Chall. 1948.
\newblock \href {https://www.jstor.org/stable/1473669} {A {Formula} for
  {Predicting} {Readability}: {Instructions}}.
\newblock \emph{Educational Research Bulletin}, 27(2):37--54.

\bibitem[{Deutsch et~al.(2020)Deutsch, Jasbi, and
  Shieber}]{deutsch_linguistic_2020}
Tovly Deutsch, Masoud Jasbi, and Stuart Shieber. 2020.
\newblock \href {https://doi.org/10.18653/v1/2020.bea-1.1} {Linguistic
  {Features} for {Readability} {Assessment}}.
\newblock In \emph{Proceedings of the {Fifteenth} {Workshop} on {Innovative}
  {Use} of {NLP} for {Building} {Educational} {Applications}}, pages 1--17,
  Seattle, WA, USA → Online. Association for Computational Linguistics.

\bibitem[{Devlin et~al.(2019)Devlin, Chang, Lee, and
  Toutanova}]{devlin_bert_2019}
Jacob Devlin, Ming-Wei Chang, Kenton Lee, and Kristina Toutanova. 2019.
\newblock \href {https://doi.org/10.48550/arXiv.1810.04805} {{BERT}:
  {Pre}-training of {Deep} {Bidirectional} {Transformers} for {Language}
  {Understanding}}.
\newblock ArXiv:1810.04805 [cs].

\bibitem[{El-Haj et~al.(2024)El-Haj, Almujaiwel, Premasiri, Ranasinghe, and
  Mitkov}]{el-haj_dares_2024}
Mo~El-Haj, Sultan Almujaiwel, Damith Premasiri, Tharindu Ranasinghe, and Ruslan
  Mitkov. 2024.
\newblock \href {https://aclanthology.org/2024.determit-1.10/} {{DARES}:
  {Dataset} for {Arabic} {Readability} {Estimation} of {School} {Materials}}.
\newblock In \emph{Proceedings of the {Workshop} on {DeTermIt}! {Evaluating}
  {Text} {Difficulty} in a {Multilingual} {Context} @ {LREC}-{COLING} 2024},
  pages 103--113, Torino, Italia. ELRA and ICCL.

\bibitem[{Flesch(1948)}]{flesch_new_1948}
Rudolph Flesch. 1948.
\newblock \href {https://doi.org/10.1037/h0057532} {A new readability
  yardstick.}
\newblock \emph{Journal of Applied Psychology}, 32(3):221--233.

\bibitem[{Gupta and Jones(2023)}]{gupta_analyzing_2023}
Esther Gupta and Douglas Jones. 2023.
\newblock \href {https://doi.org/10.21437/SLaTE.2023-16} {Analyzing the {Trade}
  {Space} in {Multi}-lingual {Automatic} {Text} {Difficulty} {Estimation}}.
\newblock In \emph{9th {Workshop} on {Speech} and {Language} {Technology} in
  {Education} ({SLaTE})}, pages 76--80. ISCA.

\bibitem[{Heilman et~al.(2008)Heilman, Collins-Thompson, and
  Eskenazi}]{heilman_analysis_2008}
Michael Heilman, Kevyn Collins-Thompson, and Maxine Eskenazi. 2008.
\newblock \href {https://aclanthology.org/W08-0909/} {An {Analysis} of
  {Statistical} {Models} and {Features} for {Reading} {Difficulty}
  {Prediction}}.
\newblock In \emph{Proceedings of the {Third} {Workshop} on {Innovative} {Use}
  of {NLP} for {Building} {Educational} {Applications}}, pages 71--79,
  Columbus, Ohio. Association for Computational Linguistics.

\bibitem[{Imperial(2021)}]{imperial_bert_2021}
Joseph~Marvin Imperial. 2021.
\newblock \href {https://aclanthology.org/2021.ranlp-1.69/} {{BERT}
  {Embeddings} for {Automatic} {Readability} {Assessment}}.
\newblock In \emph{Proceedings of the {International} {Conference} on {Recent}
  {Advances} in {Natural} {Language} {Processing} ({RANLP} 2021)}, pages
  611--618, Held Online. INCOMA Ltd.

\bibitem[{Imperial et~al.(2025)Imperial, Barayan, Stodden, Wilkens,
  Muñoz~Sánchez, Gao, Torgbi, Knight, Forey, Jablonkai, Kochmar, Reynolds,
  Ribeiro, Saggion, Volodina, Vajjala, François, Alva-Manchego, and
  Tayyar~Madabushi}]{imperial_universalcefr_2025}
Joseph~Marvin Imperial, Abdullah Barayan, Regina Stodden, Rodrigo Wilkens,
  Ricardo Muñoz~Sánchez, Lingyun Gao, Melissa Torgbi, Dawn Knight, Gail
  Forey, Reka~R. Jablonkai, Ekaterina Kochmar, Robert~Joshua Reynolds, Eugénio
  Ribeiro, Horacio Saggion, Elena Volodina, Sowmya Vajjala, Thomas François,
  Fernando Alva-Manchego, and Harish Tayyar~Madabushi. 2025.
\newblock \href {https://doi.org/10.18653/v1/2025.emnlp-main.491}
  {{UniversalCEFR}: {Enabling} {Open} {Multilingual} {Research} on {Language}
  {Proficiency} {Assessment}}.
\newblock In \emph{Proceedings of the 2025 {Conference} on {Empirical}
  {Methods} in {Natural} {Language} {Processing}}, pages 9703--9755, Suzhou,
  China. Association for Computational Linguistics.

\bibitem[{Imperial and Ong(2021)}]{imperial_under_2021}
Joseph~Marvin Imperial and Ethel Ong. 2021.
\newblock \href {https://aclanthology.org/2021.paclic-1.1/} {Under the
  {Microscope}: {Interpreting} {Readability} {Assessment} {Models} for
  {Filipino}}.
\newblock In \emph{Proceedings of the 35th {Pacific} {Asia} {Conference} on
  {Language}, {Information} and {Computation}}, pages 1--10, Shanghai, China.
  Association for Computational Lingustics.

\bibitem[{Khanuja et~al.(2021)Khanuja, Bansal, Mehtani, Khosla, Dey, Gopalan,
  Margam, Aggarwal, Nagipogu, Dave, Gupta, Gali, Subramanian, and
  Talukdar}]{khanuja_muril_2021}
Simran Khanuja, Diksha Bansal, Sarvesh Mehtani, Savya Khosla, Atreyee Dey,
  Balaji Gopalan, Dilip~Kumar Margam, Pooja Aggarwal, Rajiv~Teja Nagipogu,
  Shachi Dave, Shruti Gupta, Subhash Chandra~Bose Gali, Vish Subramanian, and
  Partha Talukdar. 2021.
\newblock \href {https://doi.org/10.48550/arXiv.2103.10730} {{MuRIL}:
  {Multilingual} {Representations} for {Indian} {Languages}}.
\newblock ArXiv:2103.10730 [cs].

\bibitem[{Kim et~al.(2018)Kim, Wattenberg, Gilmer, Cai, Wexler, Viegas, and
  Sayres}]{kim_interpretability_2018}
Been Kim, Martin Wattenberg, Justin Gilmer, Carrie Cai, James Wexler, Fernanda
  Viegas, and Rory Sayres. 2018.
\newblock \href {https://doi.org/10.48550/arXiv.1711.11279} {Interpretability
  {Beyond} {Feature} {Attribution}: {Quantitative} {Testing} with {Concept}
  {Activation} {Vectors} ({TCAV})}.
\newblock ArXiv:1711.11279 [stat].

\bibitem[{Kincaid et~al.(1975)Kincaid, Fishburne, Rogers, and
  Chissom}]{kincaid_derivation_1975}
J.~Peter Kincaid, Robert~P. Fishburne, Jr., Richard~L. Rogers, and Brad~S.
  Chissom. 1975.
\newblock \href {https://apps.dtic.mil/sti/html/tr/ADA006655/} {Derivation of
  {New} {Readability} {Formulas} ({Automated} {Readability} {Index}, {Fog}
  {Count} and {Flesch} {Reading} {Ease} {Formula}) for {Navy} {Enlisted}
  {Personnel}}.
\newblock Technical Report Research Branch Report 8-75, Naval Technical
  Training Command, Millington, TN, Research Branch.

\bibitem[{Kuratov and Arkhipov(2019)}]{kuratov_adaptation_2019}
Yuri Kuratov and Mikhail Arkhipov. 2019.
\newblock \href {https://doi.org/10.48550/arXiv.1905.07213} {Adaptation of
  {Deep} {Bidirectional} {Multilingual} {Transformers} for {Russian}
  {Language}}.
\newblock ArXiv:1905.07213 [cs].

\bibitem[{Lee et~al.(2021)Lee, Jang, and Lee}]{lee_pushing_2021}
Bruce~W. Lee, Yoo~Sung Jang, and Jason Lee. 2021.
\newblock \href {https://doi.org/10.18653/v1/2021.emnlp-main.834} {Pushing on
  {Text} {Readability} {Assessment}: {A} {Transformer} {Meets} {Handcrafted}
  {Linguistic} {Features}}.
\newblock In \emph{Proceedings of the 2021 {Conference} on {Empirical}
  {Methods} in {Natural} {Language} {Processing}}, pages 10669--10686, Online
  and Punta Cana, Dominican Republic. Association for Computational
  Linguistics.

\bibitem[{Lee and Lee(2023)}]{lee_lftk_2023}
Bruce~W. Lee and Jason Lee. 2023.
\newblock \href {https://doi.org/10.18653/v1/2023.bea-1.1} {{LFTK}:
  {Handcrafted} {Features} in {Computational} {Linguistics}}.
\newblock In \emph{Proceedings of the 18th {Workshop} on {Innovative} {Use} of
  {NLP} for {Building} {Educational} {Applications} ({BEA} 2023)}, pages 1--19,
  Toronto, Canada. Association for Computational Linguistics.

\bibitem[{Li et~al.(2024)Li, Jin, Huang, Xu, Lian, Lin, Zhang, and
  Wang}]{li_evaluating_2024}
Meng Li, Haoran Jin, Ruixuan Huang, Zhihao Xu, Defu Lian, Zijia Lin, Di~Zhang,
  and Xiting Wang. 2024.
\newblock \href {https://doi.org/10.18653/v1/2024.emnlp-main.36} {Evaluating
  {Readability} and {Faithfulness} of {Concept}-based {Explanations}}.
\newblock In \emph{Proceedings of the 2024 {Conference} on {Empirical}
  {Methods} in {Natural} {Language} {Processing}}, pages 607--625, Miami,
  Florida, USA. Association for Computational Linguistics.

\bibitem[{Liu and Lee(2023)}]{liu_hybrid_2023}
Fengkai Liu and John Lee. 2023.
\newblock \href {https://doi.org/10.18653/v1/2023.bea-1.37} {Hybrid {Models}
  for {Sentence} {Readability} {Assessment}}.
\newblock In \emph{Proceedings of the 18th {Workshop} on {Innovative} {Use} of
  {NLP} for {Building} {Educational} {Applications} ({BEA} 2023)}, pages
  448--454, Toronto, Canada. Association for Computational Linguistics.

\bibitem[{Lundberg and Lee(2017)}]{lundberg_unified_2017}
Scott~M Lundberg and Su-In Lee. 2017.
\newblock \href
  {https://papers.nips.cc/paper_files/paper/2017/hash/8a20a8621978632d76c43dfd28b67767-Abstract.html}
  {A {Unified} {Approach} to {Interpreting} {Model} {Predictions}}.
\newblock In \emph{Advances in {Neural} {Information} {Processing} {Systems}},
  volume~30. Curran Associates, Inc.

\bibitem[{Martin et~al.(2020)Martin, Muller, Ortiz~Suárez, Dupont, Romary,
  de~la Clergerie, Seddah, and Sagot}]{martin_camembert_2020}
Louis Martin, Benjamin Muller, Pedro~Javier Ortiz~Suárez, Yoann Dupont,
  Laurent Romary, Éric de~la Clergerie, Djamé Seddah, and Benoît Sagot.
  2020.
\newblock \href {https://doi.org/10.18653/v1/2020.acl-main.645} {{CamemBERT}: a
  {Tasty} {French} {Language} {Model}}.
\newblock In \emph{Proceedings of the 58th {Annual} {Meeting} of the
  {Association} for {Computational} {Linguistics}}, pages 7203--7219, Online.
  Association for Computational Linguistics.

\bibitem[{Martinc et~al.(2021)Martinc, Pollak, and
  Robnik-Šikonja}]{martinc_supervised_2021}
Matej Martinc, Senja Pollak, and Marko Robnik-Šikonja. 2021.
\newblock \href {https://doi.org/10.1162/coli_a_00398} {Supervised and
  {Unsupervised} {Neural} {Approaches} to {Text} {Readability}}.
\newblock \emph{Computational Linguistics}, 47(1):141--179.

\bibitem[{Maudslay and Cotterell(2021)}]{maudslay_syntactic_2021}
Rowan~Hall Maudslay and Ryan Cotterell. 2021.
\newblock \href {https://doi.org/10.48550/arXiv.2106.02559} {Do {Syntactic}
  {Probes} {Probe} {Syntax}? {Experiments} with {Jabberwocky} {Probing}}.
\newblock ArXiv:2106.02559 [cs].

\bibitem[{Mc~Laughlin(1969)}]{mc_laughlin_smog_1969}
G.~Harry Mc~Laughlin. 1969.
\newblock \href {https://www.jstor.org/stable/40011226} {{SMOG} {Grading}-a
  {New} {Readability} {Formula}}.
\newblock \emph{Journal of Reading}, 12(8):639--646.

\bibitem[{Naous et~al.(2024)Naous, Ryan, Lavrouk, Chandra, and
  Xu}]{naous_readme_2024}
Tarek Naous, Michael~J Ryan, Anton Lavrouk, Mohit Chandra, and Wei Xu. 2024.
\newblock \href {https://doi.org/10.18653/v1/2024.emnlp-main.682} {{ReadMe}++:
  {Benchmarking} {Multilingual} {Language} {Models} for {Multi}-{Domain}
  {Readability} {Assessment}}.
\newblock In \emph{Proceedings of the 2024 {Conference} on {Empirical}
  {Methods} in {Natural} {Language} {Processing}}, pages 12230--12266, Miami,
  Florida, USA. Association for Computational Linguistics.

\bibitem[{Nejadgholi et~al.(2022{\natexlab{a}})Nejadgholi, Balkir, Fraser, and
  Kiritchenko}]{nejadgholi_towards_2022}
Isar Nejadgholi, Esma Balkir, Kathleen Fraser, and Svetlana Kiritchenko.
  2022{\natexlab{a}}.
\newblock \href {https://doi.org/10.18653/v1/2022.blackboxnlp-1.18} {Towards
  {Procedural} {Fairness}: {Uncovering} {Biases} in {How} a {Toxic} {Language}
  {Classifier} {Uses} {Sentiment} {Information}}.
\newblock In \emph{Proceedings of the {Fifth} {BlackboxNLP} {Workshop} on
  {Analyzing} and {Interpreting} {Neural} {Networks} for {NLP}}, pages
  225--237, Abu Dhabi, United Arab Emirates (Hybrid). Association for
  Computational Linguistics.

\bibitem[{Nejadgholi et~al.(2022{\natexlab{b}})Nejadgholi, Fraser, and
  Kiritchenko}]{nejadgholi_improving_2022}
Isar Nejadgholi, Kathleen Fraser, and Svetlana Kiritchenko. 2022{\natexlab{b}}.
\newblock \href {https://doi.org/10.18653/v1/2022.acl-long.378} {Improving
  {Generalizability} in {Implicitly} {Abusive} {Language} {Detection} with
  {Concept} {Activation} {Vectors}}.
\newblock In \emph{Proceedings of the 60th {Annual} {Meeting} of the
  {Association} for {Computational} {Linguistics} ({Volume} 1: {Long}
  {Papers})}, pages 5517--5529, Dublin, Ireland. Association for Computational
  Linguistics.

\bibitem[{Nejadgholi et~al.(2023)Nejadgholi, Kiritchenko, Fraser, and
  Balkir}]{nejadgholi_concept-based_2023}
Isar Nejadgholi, Svetlana Kiritchenko, Kathleen~C. Fraser, and Esma Balkir.
  2023.
\newblock \href {https://doi.org/10.18653/v1/2023.woah-1.14} {Concept-{Based}
  {Explanations} to {Test} for {False} {Causal} {Relationships} {Learned} by
  {Abusive} {Language} {Classifiers}}.
\newblock In \emph{The 7th {Workshop} on {Online} {Abuse} and {Harms}
  ({WOAH})}, pages 138--149, Toronto, Canada. Association for Computational
  Linguistics.

\bibitem[{Rao et~al.(2021)Rao, Zheng, and Li}]{rao_cross-lingual_2021}
Simin Rao, Hua Zheng, and Sujian Li. 2021.
\newblock \href {https://doi.org/10.18653/v1/2021.findings-emnlp.227}
  {Cross-{Lingual} {Leveled} {Reading} {Based} on {Language}-{Invariant}
  {Features}}.
\newblock In \emph{Findings of the {Association} for {Computational}
  {Linguistics}: {EMNLP} 2021}, pages 2677--2682, Punta Cana, Dominican
  Republic. Association for Computational Linguistics.

\bibitem[{Ribeiro-Flucht et~al.(2024)Ribeiro-Flucht, Chen, and
  Meurers}]{ribeiro-flucht_explainable_2024}
Luisa Ribeiro-Flucht, Xiaobin Chen, and Detmar Meurers. 2024.
\newblock \href {https://aclanthology.org/2024.bea-1.17/} {Explainable {AI} in
  {Language} {Learning}: {Linking} {Empirical} {Evidence} and {Theoretical}
  {Concepts} in {Proficiency} and {Readability} {Modeling} of {Portuguese}}.
\newblock In \emph{Proceedings of the 19th {Workshop} on {Innovative} {Use} of
  {NLP} for {Building} {Educational} {Applications} ({BEA} 2024)}, pages
  199--209, Mexico City, Mexico. Association for Computational Linguistics.

\bibitem[{Sarti et~al.(2021)Sarti, Brunato, and Dell'Orletta}]{sarti_that_2021}
Gabriele Sarti, Dominique Brunato, and Felice Dell'Orletta. 2021.
\newblock \href {https://doi.org/10.18653/v1/2021.cmcl-1.5} {That {Looks}
  {Hard}: {Characterizing} {Linguistic} {Complexity} in {Humans} and {Language}
  {Models}}.
\newblock In \emph{Proceedings of the {Workshop} on {Cognitive} {Modeling} and
  {Computational} {Linguistics}}, pages 48--60, Online. Association for
  Computational Linguistics.

\bibitem[{Shen et~al.(2013)Shen, Williams, Marius, and
  Salesky}]{shen_language-independent_2013}
Wade Shen, Jennifer Williams, Tamas Marius, and Elizabeth Salesky. 2013.
\newblock \href {https://aclanthology.org/W13-2904/} {A
  {Language}-{Independent} {Approach} to {Automatic} {Text} {Difficulty}
  {Assessment} for {Second}-{Language} {Learners}}.
\newblock In \emph{Proceedings of the {Second} {Workshop} on {Predicting} and
  {Improving} {Text} {Readability} for {Target} {Reader} {Populations}}, pages
  30--38, Sofia, Bulgaria. Association for Computational Linguistics.

\bibitem[{Tenney et~al.(2019)Tenney, Das, and Pavlick}]{tenney_bert_2019}
Ian Tenney, Dipanjan Das, and Ellie Pavlick. 2019.
\newblock \href {https://doi.org/10.48550/arXiv.1905.05950} {{BERT}
  {Rediscovers} the {Classical} {NLP} {Pipeline}}.
\newblock ArXiv:1905.05950 [cs].

\bibitem[{Vajjala(2022)}]{vajjala_trends_2022}
Sowmya Vajjala. 2022.
\newblock \href {https://aclanthology.org/2022.lrec-1.574/} {Trends,
  {Limitations} and {Open} {Challenges} in {Automatic} {Readability}
  {Assessment} {Research}}.
\newblock In \emph{Proceedings of the {Thirteenth} {Language} {Resources} and
  {Evaluation} {Conference}}, pages 5366--5377, Marseille, France. European
  Language Resources Association.

\bibitem[{Vajjala and Meurers(2013)}]{vajjala_applicability_2013}
Sowmya Vajjala and Detmar Meurers. 2013.
\newblock \href {https://aclanthology.org/W13-2907/} {On {The} {Applicability}
  of {Readability} {Models} to {Web} {Texts}}.
\newblock In \emph{Proceedings of the {Second} {Workshop} on {Predicting} and
  {Improving} {Text} {Readability} for {Target} {Reader} {Populations}}, pages
  59--68, Sofia, Bulgaria. Association for Computational Linguistics.

\bibitem[{Wilkens et~al.(2024{\natexlab{a}})Wilkens, Watrin, Cardon, Pintard,
  Gribomont, and François}]{wilkens_exploring_2024}
Rodrigo Wilkens, Patrick Watrin, Rémi Cardon, Alice Pintard, Isabelle
  Gribomont, and Thomas François. 2024{\natexlab{a}}.
\newblock \href {https://aclanthology.org/2024.findings-eacl.153/} {Exploring
  hybrid approaches to readability: experiments on the complementarity between
  linguistic features and transformers}.
\newblock In \emph{Findings of the {Association} for {Computational}
  {Linguistics}: {EACL} 2024}, pages 2316--2331, St. Julian's, Malta.
  Association for Computational Linguistics.

\bibitem[{Wilkens et~al.(2024{\natexlab{b}})Wilkens, Watrin, and
  François}]{wilkens_paying_2024}
Rodrigo Wilkens, Patrick Watrin, and Thomas François. 2024{\natexlab{b}}.
\newblock \href {https://aclanthology.org/2024.readi-1.9/} {Paying attention to
  the words: explaining readability prediction for {French} as a foreign
  language}.
\newblock In \emph{Proceedings of the 3rd {Workshop} on {Tools} and {Resources}
  for {People} with {REAding} {DIfficulties} ({READI}) @ {LREC}-{COLING} 2024},
  pages 102--115, Torino, Italia. ELRA and ICCL.

\bibitem[{Zhao et~al.(2023)Zhao, Chen, Yang, Liu, Deng, Cai, Wang, Yin, and
  Du}]{zhao_explainability_2023}
Haiyan Zhao, Hanjie Chen, Fan Yang, Ninghao Liu, Huiqi Deng, Hengyi Cai,
  Shuaiqiang Wang, Dawei Yin, and Mengnan Du. 2023.
\newblock \href {https://doi.org/10.48550/arXiv.2309.01029} {Explainability for
  {Large} {Language} {Models}: {A} {Survey}}.
\newblock ArXiv:2309.01029 [cs].

\end{thebibliography}
\bibliographystyle{acl_natbib}

\appendix

\section{Additional SHAP Results}
\label{app:shap}

\begin{figure}[h]
    \centering
    \includegraphics[width=\columnwidth]{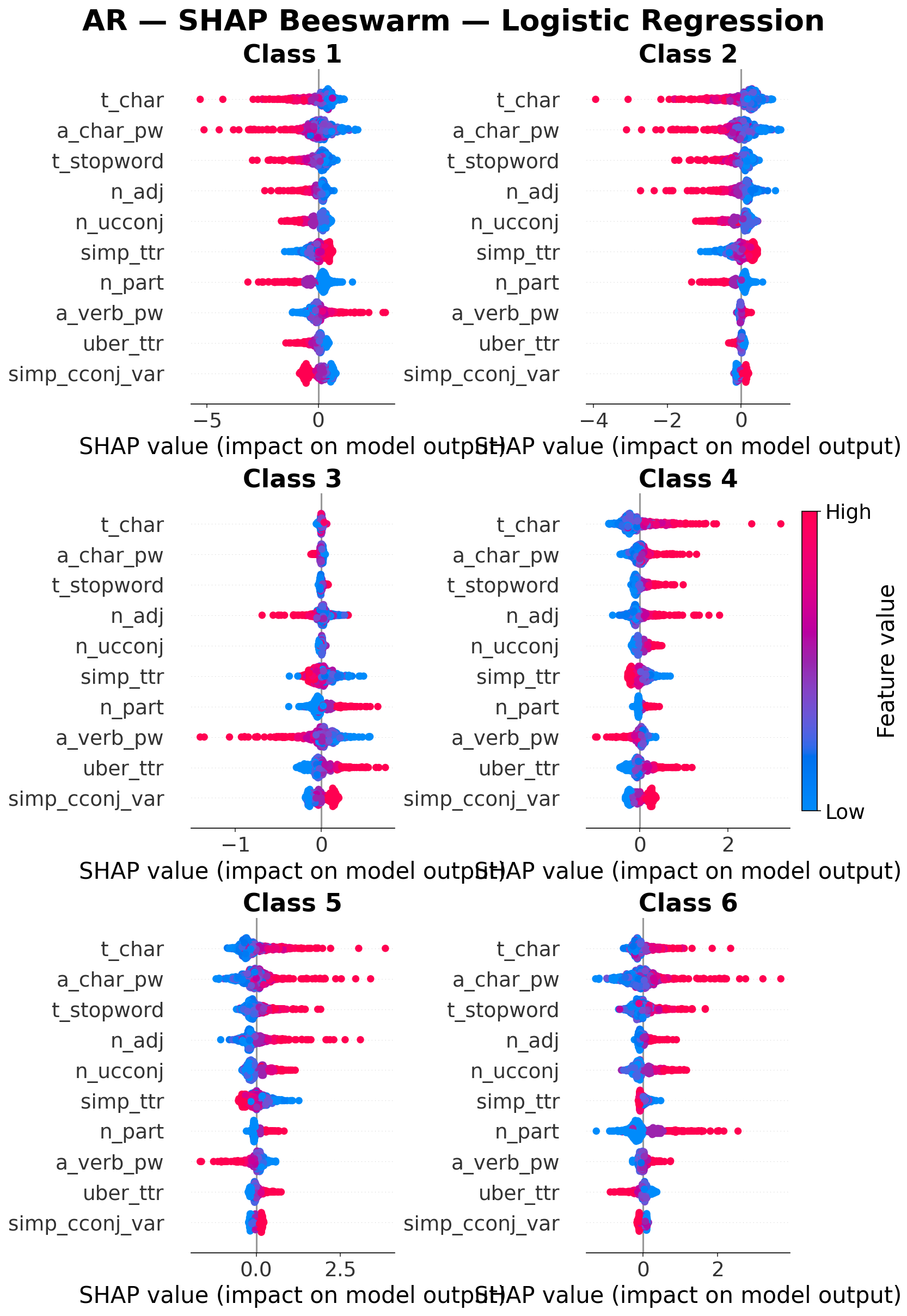}
    \caption{Arabic Logistic Regression SHAP beeswarm, one panel per CEFR level.}
    \label{fig:shap_ar}
\end{figure}

\begin{figure}[h]
    \centering
    \includegraphics[width=\columnwidth]{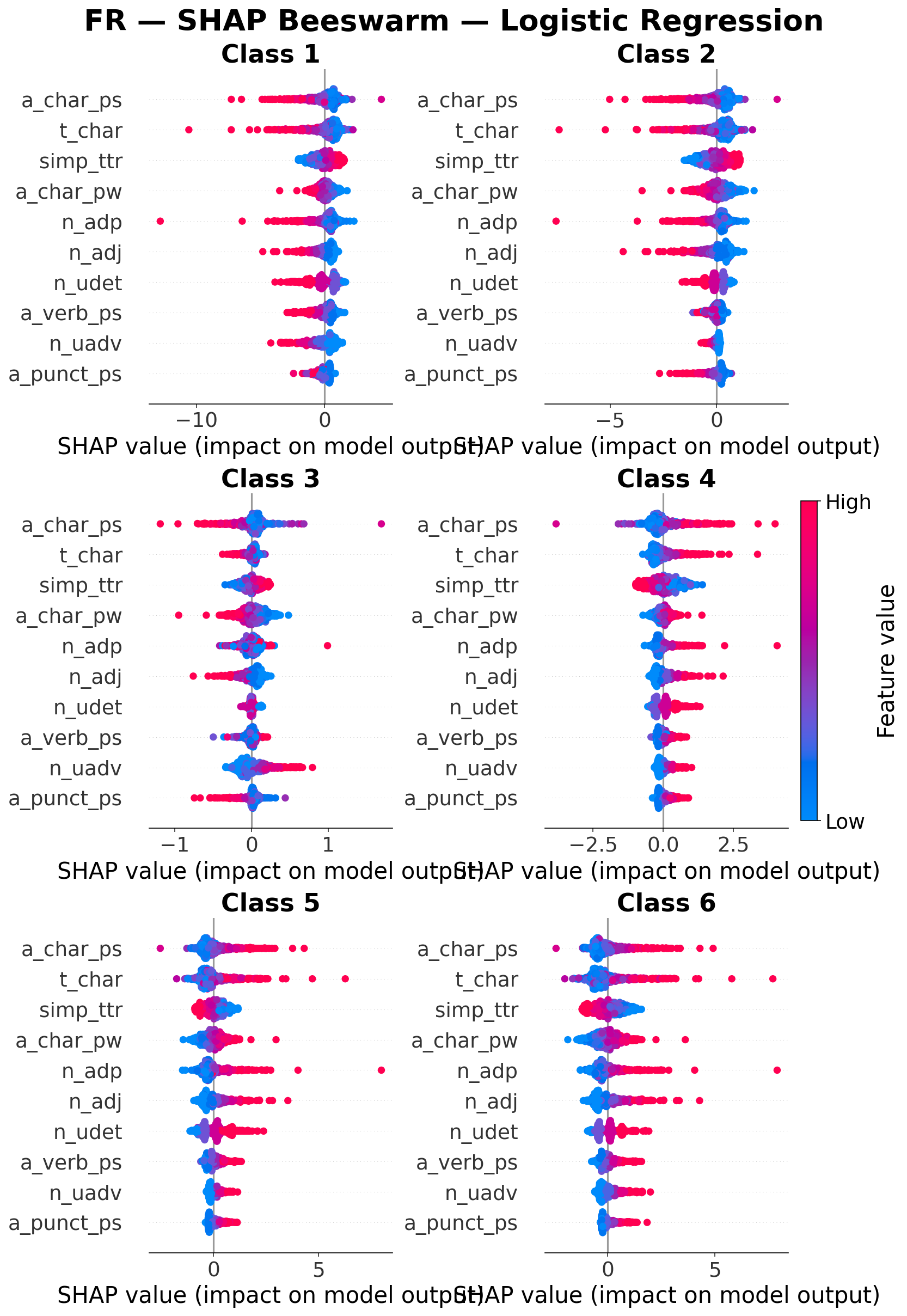}
    \caption{French Logistic Regression SHAP beeswarm, one panel per CEFR level.}
    \label{fig:shap_fr}
\end{figure}

\begin{figure}[h]
    \centering
    \includegraphics[width=\columnwidth]{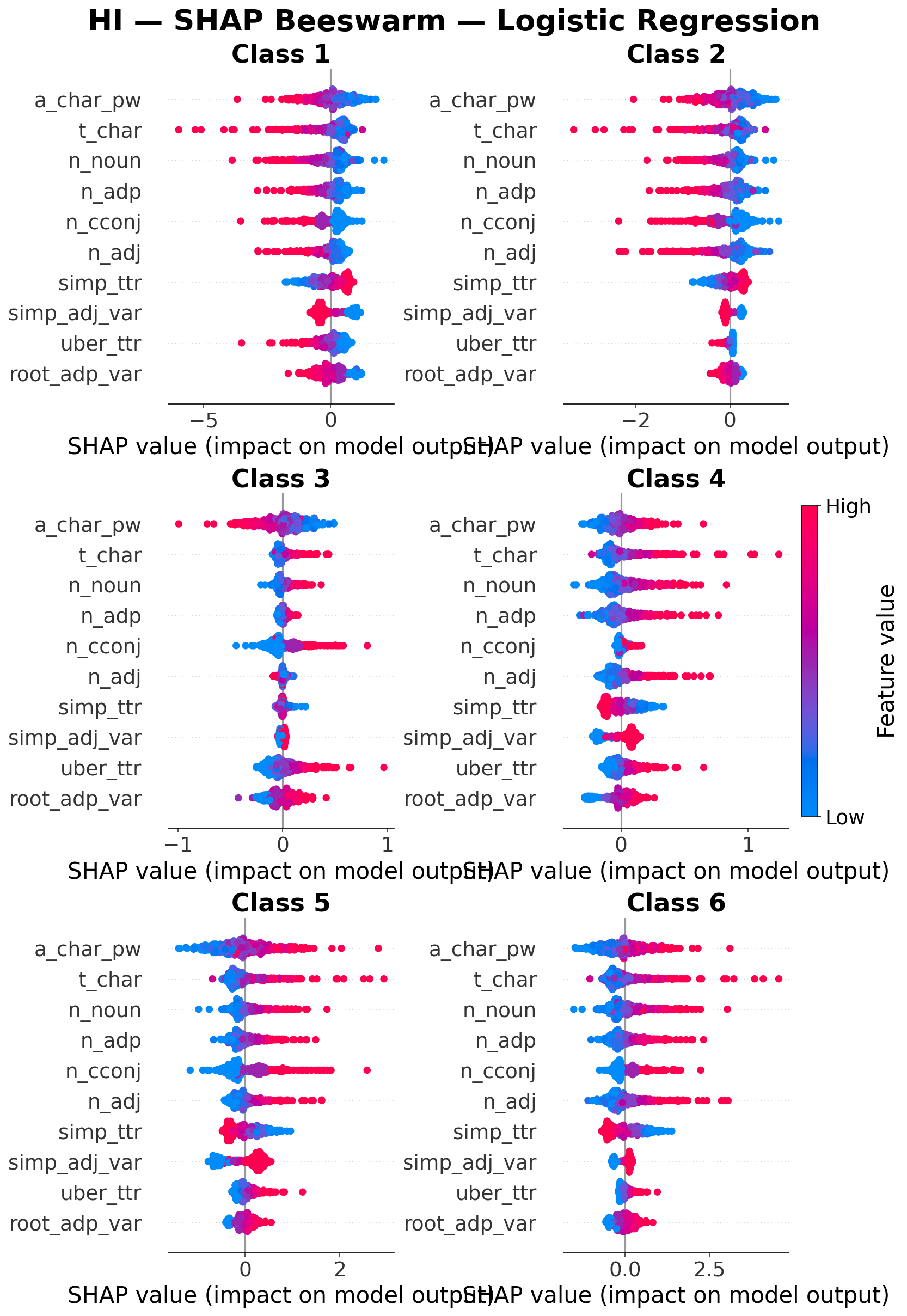}
    \caption{Hindi Logistic Regression SHAP beeswarm, one panel per CEFR level.}
    \label{fig:shap_hi}
\end{figure}

\begin{figure}[h]
    \centering
    \includegraphics[width=\columnwidth]{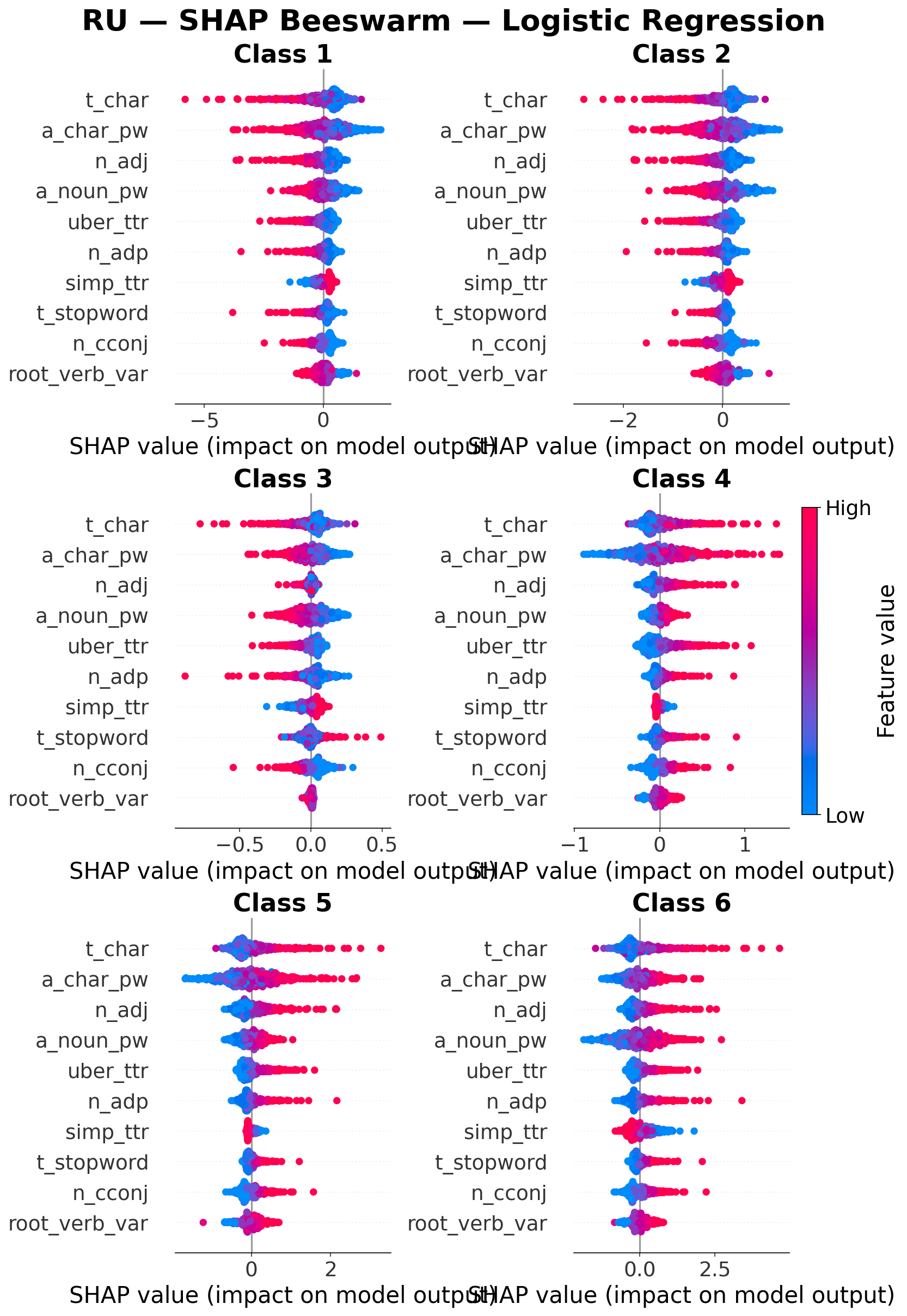}
    \caption{Russian Logistic Regression SHAP beeswarm, one panel per CEFR level.}
    \label{fig:shap_ru}
\end{figure}

Figures~\ref{fig:shap_ar}--\ref{fig:shap_ru} show SHAP beeswarms from the Logistic Regression classifier for the four languages not shown in the main text. Each panel corresponds to one CEFR class (Class 0 = A1, Class 5 = C2), with color encoding feature value (blue = low, red = high) and horizontal position encoding SHAP contribution. The ordinal polarity between A1 and C2 reported in the main text holds across all languages.

\section{Additional TCAV Results}
\label{app:tcav}

\begin{figure}[h]
    \centering
    \includegraphics[width=\columnwidth]{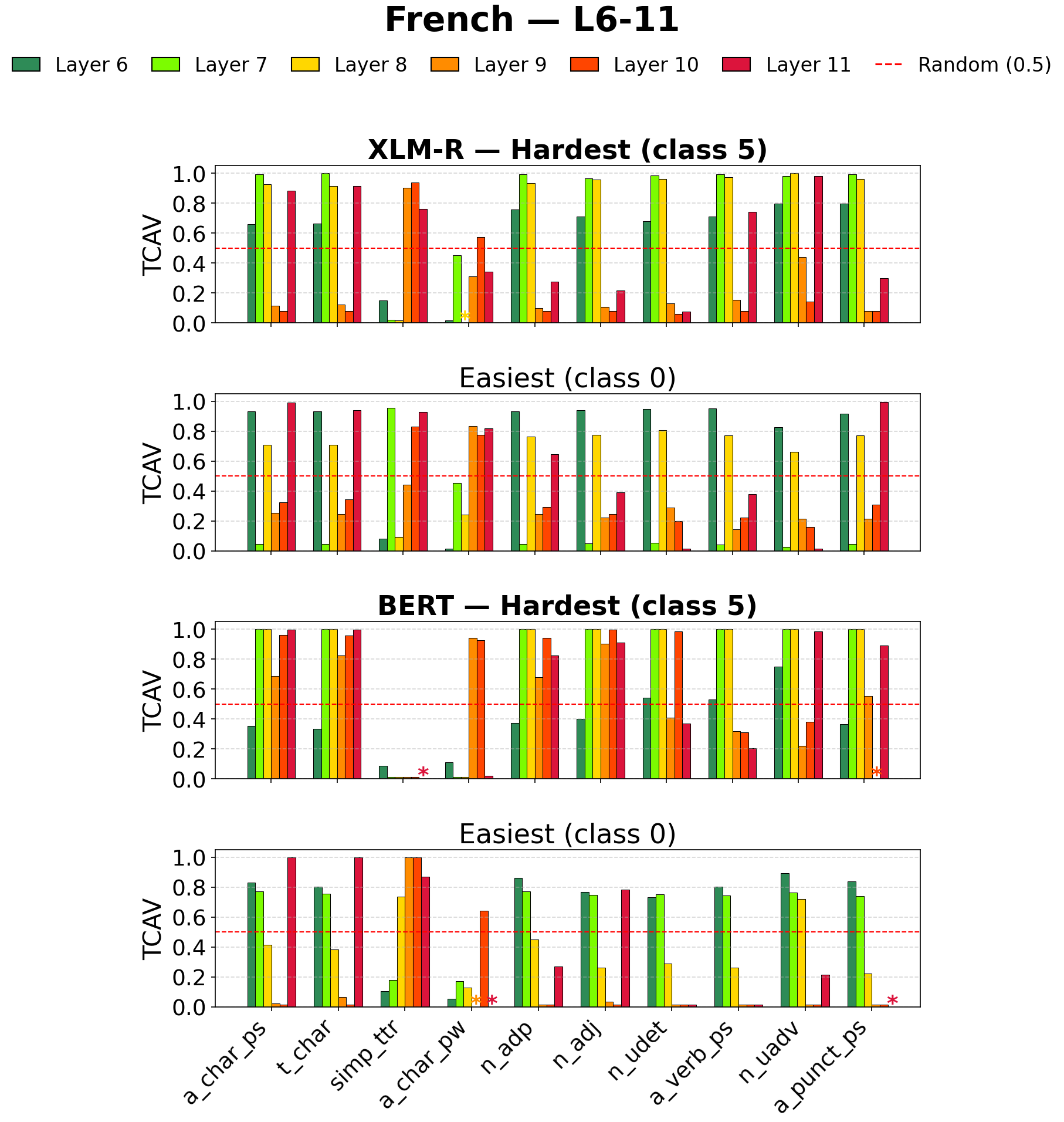}
    \caption{French TCAV fractions for the top 10 SHAP features, layers 6--11. Top half: XLM-R. Bottom half: CamemBERT. For each model, the upper panel shows C2 and the lower panel shows A1.}
    \label{fig:tcav_fr}
\end{figure}

\begin{figure}[h]
    \centering
    \includegraphics[width=\columnwidth]{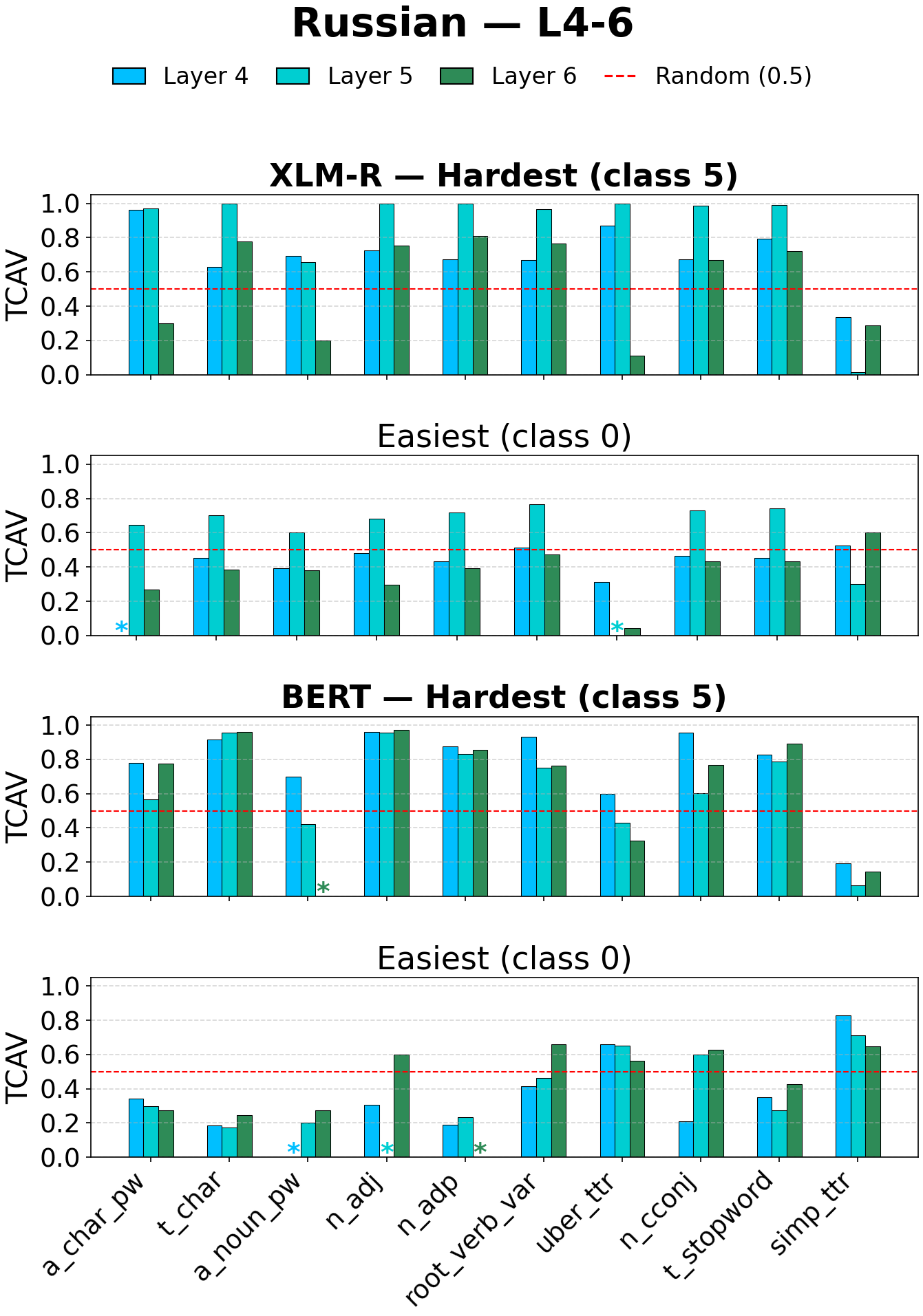}
    \caption{Russian TCAV fractions for the top 10 SHAP features, layers 4--6. Top half: XLM-R. Bottom half: RuBERT. For each model, the upper panel shows C2 and the lower panel shows A1.}
    \label{fig:tcav_ru}
\end{figure}

Figures~\ref{fig:tcav_fr} and \ref{fig:tcav_ru} show TCAV fractions for the top 10 SHAP features in French and Russian, across the mid-to-late layer groups with a coherent directional pattern. Each figure shows XLM-R in the top half and the language-specific encoder in the bottom. Within each half, the upper panel shows C2 (hardest) and the lower panel shows A1 (easiest). Asterisks mark non-significant results.

\section{Additional CAV Accuracy Results}
\label{app:cav_accuracy}

\begin{figure}[h]
    \centering
    \includegraphics[width=\columnwidth]{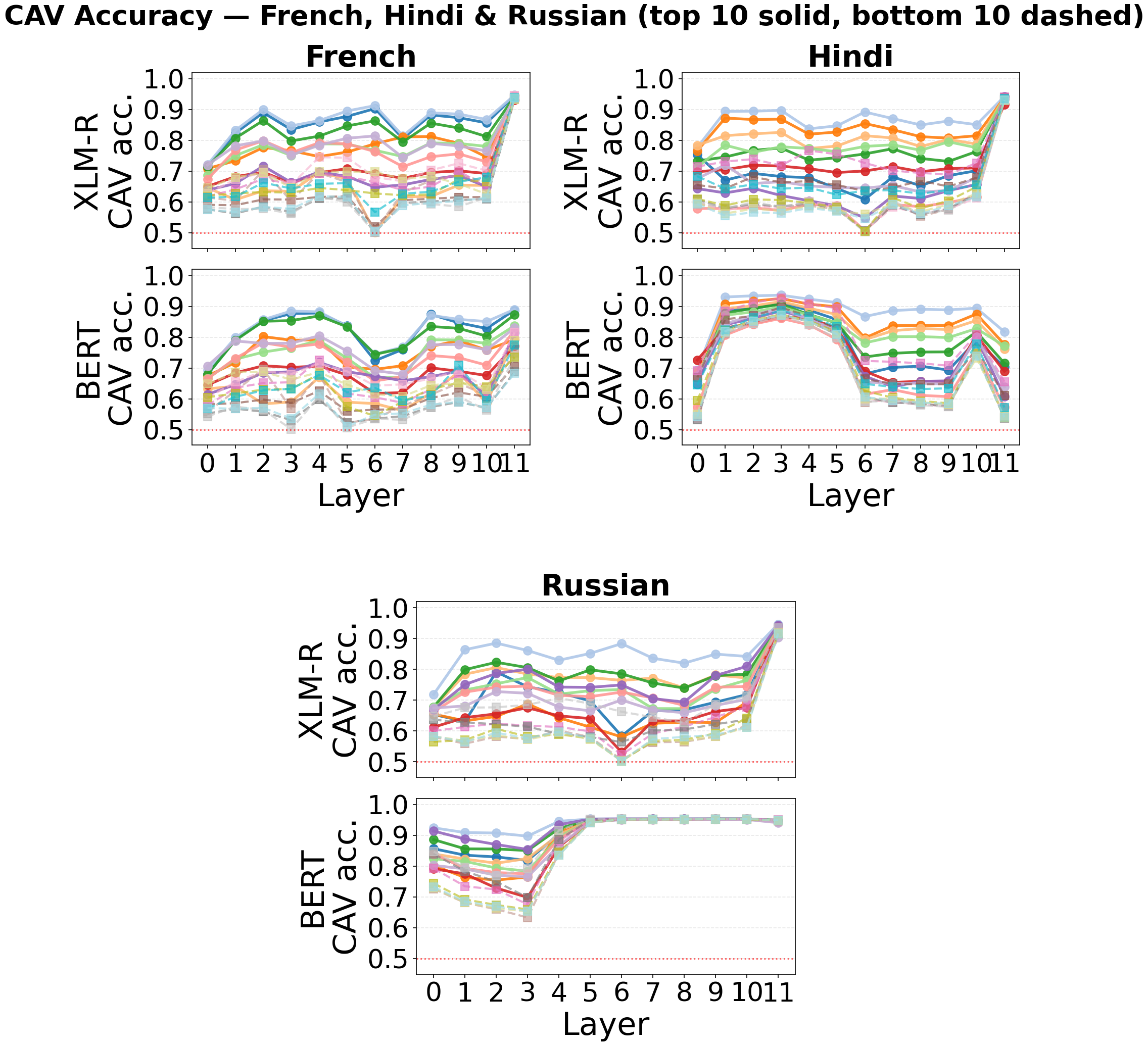}
    \caption{Mean CAV classifier accuracy across 12 layers for French, Hindi, and Russian. Top row: XLM-R. Bottom row: language-specific encoder. Solid lines show top 10 SHAP features; dashed lines show bottom 10.}
    \label{fig:cav_acc_appendix}
\end{figure}

Figure~\ref{fig:cav_acc_appendix} shows mean CAV classifier accuracy across 12 layers for French, Hindi, and Russian, complementing the Arabic and English plots in the main text. The same gap between top 10 and bottom 10 SHAP features holds across these languages.

\section{Use of AI Assistants}
\label{app:ai_assistants}

We used AI assistants during this work, primarily Claude (including Claude Code) and ChatGPT. For writing, AI assistants were used to edit and polish grammar and for LaTeX formatting. For coding, Claude Code was used to assist in writing and iterating on the Python code for our experiments. All experimental design, choice of methodology, concept formulation (SHAP--TCAV framework, feature selection, probing setup), interpretation of results, and the scientific content of the paper are the authors' own. The authors reviewed, tested, and verified AI-assisted code and text before inclusion, and take full responsibility for the correctness of the experiments and claims in this paper.

\end{document}